\documentclass[10pt,twocolumn,letterpaper]{article}
\usepackage{silence}
\usepackage{cvpr}      %

\usepackage{bm}
\usepackage{balance}
\usepackage{comment}
\usepackage{amsmath}
\DeclareMathOperator*{\argmin}{argmin}

\usepackage[dvipsnames]{xcolor}

\definecolor{cvprblue}{rgb}{0.21,0.49,0.74}
\usepackage[pagebackref,breaklinks,colorlinks,citecolor=cvprblue]{hyperref}

\title{{TECA}: Text-Guided Generation and Editing of Compositional {3D} Avatars}

\author{\fontsize{11.5pt}{\baselineskip}\selectfont Hao Zhang\textsuperscript{1,3,4*},~~Yao Feng\textsuperscript{1,2*},~~ Peter Kulits\textsuperscript{1},~~Yandong Wen\textsuperscript{1}\\\fontsize{11.5pt}{\baselineskip}\selectfont Justus Thies \textsuperscript{1},~~Michael J. Black\textsuperscript{1}\\[0.25mm]
\fontsize{11pt}{\baselineskip}\selectfont\textsuperscript{1}Max Planck Institute for Intelligent Systems,~~\textsuperscript{2}ETH Zürich,~~\textsuperscript{3}Tsinghua University,
\\\fontsize{11pt}{\baselineskip}\selectfont\textsuperscript{4}RWTH Aachen University,~~\textsuperscript{*}Equal contribution
}

\newcommand{\tdv}[1]{\textcolor{black}{#1}}

\newcommand{\modelname}{TECA\xspace}

\newcommand{\qheading}[1]{\vspace{-1em}\paragraph{#1}}

\newcommand{\supmat}{Sup.~Mat.\xspace}

\begin{document}
\twocolumn[{
\renewcommand\twocolumn[1][]{#1}
\maketitle

\vspace{-2.5em}
\begin{center}
\vbox{%
         \vskip 0.05in
         \hsize\textwidth
         \linewidth\hsize
         \centering
         \normalsize
         {Project Page:}~\tt\href{https://yfeng95.github.io/teca}{yfeng95.github.io/teca}
         \vskip 0.05in
}
\end{center}

\begin{center}
    \captionsetup{type=figure}
    \includegraphics[width=\textwidth]{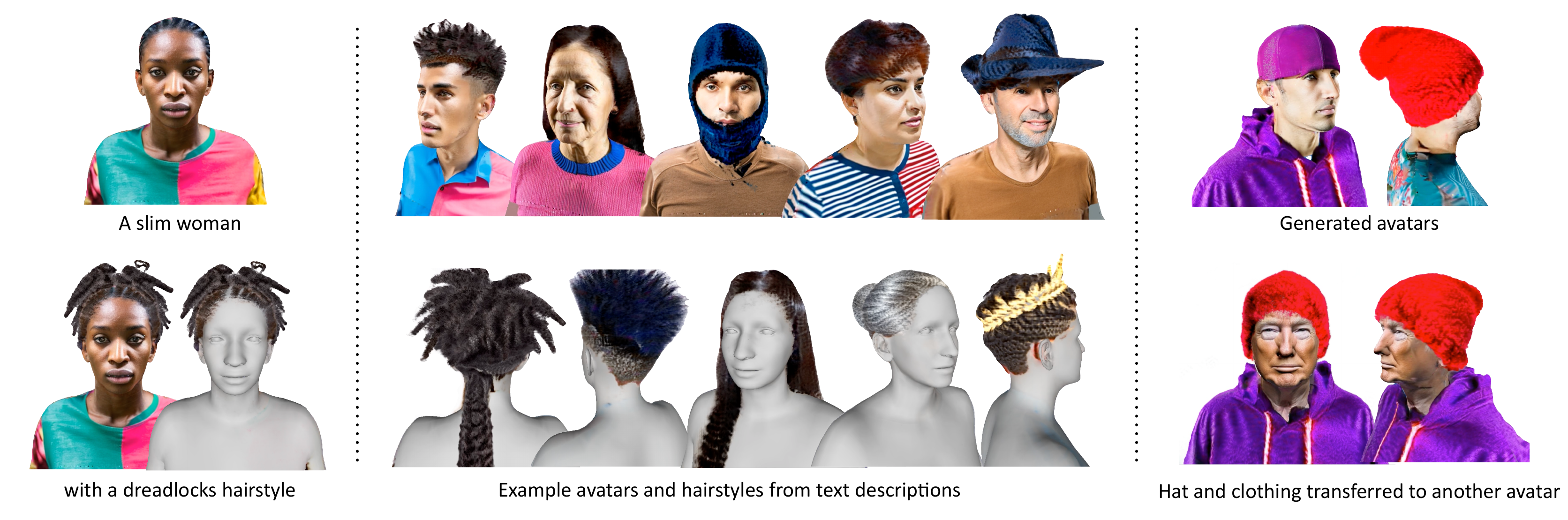}
    \captionof{figure}{Compositional avatars: (left) Given a text description, our method produces a 3D avatar consisting of a mesh-based face and body (gray) and NeRF-based style components (e.g. hair and clothing). (middle) Our method generates diverse avatars and hairstyles based on the text input. (right) The non-face parts can be seamlessly transferred to new avatars with different shapes, without additional processing.}
    \label{fig:teaser}
\end{center}
}
]

\newcommand{\vect}[1]{\mathbf{#1}}
\newcommand{\norm}[1]{\left\lVert#1\right\rVert}

\newcommand{\shapecoeff}{\boldsymbol{\beta}}
\newcommand{\shapedim}{{\left| \shapecoeff \right|}}
\newcommand{\shapespace}{\mathcal{S}}
\newcommand{\shapespaceexpl}{\mathbb{R}^{\shapedim}}

\newcommand{\numjoints}{n_k}
\newcommand{\joints}{\textbf{J}}
\newcommand{\jointregressor}{J}

\newcommand{\posecoeff}{\boldsymbol{\theta}}
\newcommand{\posedim}{{3\numjoints+3}}
\newcommand{\posespace}{\mathcal{P}}
\newcommand{\posespaceexpl}{\mathbb{R}^{\posedim}}

\newcommand{\expcoeff}{\boldsymbol{\psi}}
\newcommand{\expdim}{{\left| \expcoeff \right|}}
\newcommand{\expspace}{\mathcal{E}}
\newcommand{\expspaceexpl}{\mathbb{R}^{\expdim}}

\newcommand{\canonicalvertex}{t^c}
\newcommand{\numverts}{n_v}
\newcommand{\numfaces}{n_t}
\newcommand{\template}{\boldsymbol{T}}
\newcommand{\feature}{z}
\newcommand{\blendweight}{w}
\newcommand{\blendweights}{\boldsymbol{W}}
\newcommand{\blendweightsdim}{\numjoints \times \numverts}
\newcommand{\lbs}{F_{\text{lbs}}}
\newcommand{\nerfmodel}{F_{\text{nerf}}}
\newcommand{\deformfunc}{T_P}
\newcommand{\blendfunc}{B}

\newcommand{\verts}{\boldsymbol{V}}
\newcommand{\faces}{\mathbf{F}}
\newcommand{\vsmplx}{F_{\text{smplx}}}
\newcommand{\T}{T}
\newcommand{\matsmplx}{\mathbf{M}}
\newcommand{\offsets}{\mathbf{O}}
\newcommand{\offset}{\vect{o}}

\newcommand{\offsetmodel}{{F_{d}}}
\newcommand{\texmodel}{{F_{t}}}
\newcommand{\clothmodel}{{F_{c}}}
\newcommand{\defmodel}{{F_{m}}}

\newcommand{\inputtext}{\boldsymbol{s}}
\newcommand{\image}{I}
\newcommand{\imagemask}{S} %
\newcommand{\numframes}{n_f}
\newcommand{\framenum}{f}
\newcommand{\point}{\vect{p}}
\newcommand{\meshrender}{\mathcal{R}_m}
\newcommand{\lossweight}[1]{\lambda_{#1}}

\newcommand{\volrender}{\mathcal{R}_v}
\newcommand{\col}{\vect{c}}
\newcommand{\rayrender}{C}
\newcommand{\density}{\sigma}
\newcommand{\ray}{R}

\newcommand{\numsamples}{n}

\newcommand{\landmark}{\textbf{k}}

\newcommand{\albedo}{A}
\newcommand{\albedocoeffs}{\boldsymbol{\alpha}}

\newcommand{\albedodim}{\left| \albedocoeffs \right|}
\newcommand{\normalcoeffs}{\boldsymbol{\nu}}
\newcommand{\normaldim}{\left| \normalcoeffs \right|}

\newcommand{\uvsize}{d}

\newcommand{\twoD}{2D\xspace}
\newcommand{\threeD}{3D\xspace}
\newcommand{\sixD}{6D\xspace}

\begin{abstract}

Our goal is to create a realistic 3D facial avatar with hair and accessories using only a text description.
While this challenge has attracted significant recent interest, existing methods either lack realism, produce unrealistic shapes, or do not support editing, such as modifications to the hairstyle. 
We argue that existing methods are limited because they employ a monolithic modeling approach, using a single representation for the head, face, hair, and accessories.
Our observation is that the hair and face, for example, have very different structural qualities that benefit from different representations.
Building on this insight, we generate avatars with a compositional model, in which the head, face, and upper body are represented with traditional \threeD meshes, and the hair, clothing, and accessories with neural radiance fields (NeRF).  
The model-based mesh representation provides a strong geometric prior for the face region, improving realism while enabling editing of the person's appearance. 
By using NeRFs to represent the remaining components, our method is able to model and synthesize parts with complex geometry and appearance, such as curly hair and fluffy scarves. 
Our novel system synthesizes these high-quality compositional avatars from text descriptions.
Specifically, we generate a face image using text, fit a parametric shape model to it, and inpaint texture using diffusion models.
Conditioned on the generated face, we sequentially generate style components such as hair or clothing using Score Distillation Sampling (SDS) with guidance from CLIPSeg segmentations. 
However, this alone is not sufficient to produce avatars with a high degree of realism.
Consequently, we introduce a hierarchical approach to refine the non-face regions using a BLIP-based loss combined with SDS.
The experimental results demonstrate that our method, Text-guided generation and Editing of Compositional Avatars (\modelname), produces avatars that are more realistic than those of recent methods while being editable because of their compositional nature. 
For example, our \modelname  enables the seamless transfer of compositional features like hairstyles, scarves, and other accessories between avatars. 
This capability supports applications such as virtual try-on.
The code and generated avatars will be publicly available for research purposes at 
\tt\textit{\href{https://yfeng95.github.io/teca}{yfeng95.github.io/teca}}.

\end{abstract}

\vspace{-1em}
\section{Introduction}
There are two traditional approaches to creating facial avatars for games and social media.
The first method allows users to select attributes such as skin color, hairstyle, and accessories manually through a graphical interface.
While one can create nearly photo-realistic avatars with tools such as MetaHuman~\cite{metahuman}, the manual process is cumbersome and it is difficult to make an avatar resemble a specific individual's appearance.
The second method involves estimating facial shape and appearance from an image~\cite{tewari2017mofa,feng2021learning,zielonka2022towards,Chan2021eg3d,feng2018prn,Liu2022SCR} or
a video~\cite{grassal2022neural,zheng2022avatar,zielonka2022instant,Feng2023DELTA}. 
These methods, however, do not support editing and customization of the captured avatar.
Recently, a third approach has emerged that exploits text descriptions, generative models, and neural radiance fields~\cite{poole2023dreamfusion,lin2022magic3d,metzer2022latent,cao2023dreamavatar}.
These methods promise the easy creation of diverse and realistic avatars but suffer in terms of \threeD realism and editability.
To address these challenges, some methods~\cite{zhang2023dreamface,aneja2022clipface} incorporate additional priors from existing face or body models~\cite{SMPL:2015, FLAME:SiggraphAsia2017} for face generation and animation.
However, their ability to synthesize complex geometries like those of hair and scarves is constrained by the fixed topology of typical \threeD mesh models.

Going beyond prior work, our goal is to make text a viable interface for creating and editing realistic \threeD face avatars with accessories and hair.
Our approach is guided by two key observations:
1) Different components of the avatar, such as the hair and face, have unique geometric and appearance properties that benefit from distinct representations.
2) Statistical shape models of head and body shapes can provide valuable guidance 
to generative image models.
To exploit the first observation, we adopt a {\em compositional} approach to avatar generation, leveraging the strengths of neural and mesh-based \threeD content creation methods.
Specifically, we model an avatar as a combination of two primary components: the face/body and non-face/body regions.
To exploit the second observation, we use the SMPL-X body model~\cite{Pavlakos2019_smplifyx} to represent the shape of the head and shoulders.
By leveraging a model-based representation of face/body shape, we remove the need to model shapes. 
Instead, we focus such models on creating realistic face texture.
Exploiting 3D shapes enables us to generate realistic faces by inpainting textures with existing, pre-trained, diffusion models. 
Moreover, the integration of the shape model enables flexible body shape modifications by manipulating the parametric shape representation, facilitating the transfer of hairstyles and other accessories between avatars with different proportions.
For the non-face components like hair, clothing, and accessories, we model their shape and appearance with NeRF~\cite{mildenhall2020nerf} since it can represent diverse geometry and reflectance.  %

Starting with a textual description, our avatar generation process involves multiple steps (see Fig.\ \ref{fig:pipeline} for an illustration).
First, we use %
stable diffusion model~\cite{rombach2022high} to generate an image of a face conditioned on the text description.
Second, we optimize the shape parameters of the SMPL-X body model~\cite{Pavlakos2019_smplifyx} to obtain a shape representation of the person.
Third, inspired by TEXTure~\cite{richardson2023texture}, we leverage the estimated \threeD face shape to rotate the head and use Stable Diffusion to generate missing texture.
Fourth, and most important, we employ a sequential, compositional approach to generate additional style components.
Conditioned on the face mesh, we learn a NeRF-based component model from the text description. %
To enable the transfer of non-face components between avatars, we define a canonical space in which we use a template face shape and train the NeRF component on top of it.
Using the hybrid volume rendering technique from SCARF~\cite{feng2022capturing} and DELTA~\cite{Feng2023DELTA}, including shape skinning, we render our hybrid avatar model into the observation image space.
We then apply Score Distillation Sampling (SDS)~\cite{poole2023dreamfusion} to optimize the NeRF model.
The SDS loss provides gradients for updating NeRF to guide 2D renderings to match the text input.
While NeRF is a highly flexible method, we argue that it is not needed to represent the face/body.
Instead, narrow its focus to modeling specific components of the avatar that are not well represented by parametric face/body models, such as hair or accessories.
To that end, we use segmentation to steer the generation. For example, in the case of hair, we compute a hair segmentation mask and use it to focus NeRF on representing the hair region.
The segmentation mask is obtained by running CLIPSeg~\cite{lueddecke22_cvpr} on the current rendering and is updated iteratively throughout the generation process. 
To enhance learning efficiency, we adopt the Latent-NeRF~\cite{metzer2022latent} approach and train the NeRF model in latent space.
Finally, we refine the non-face regions using a combination of a BLIP-based loss~\cite{li2022blip} and SDS in image space. This improves the visual quality of the non-face components.

Figure \ref{fig:teaser} shows several avatars generated by our method.
It shows the underlying body mesh, the generated texture from different views, the generation of NeRF-based hair, hats, and clothing, as well as the transfer of components from one avatar to another.
Our approach for avatar generation surpasses existing methods in terms of realism, as demonstrated by extensive qualitative analysis.
\modelname' compositional framework has two key advantages.
First, it uses the ``right'' models for the task: meshes for the face and body and NeRF for hair and clothing.
This disentangling of the face and non-face parts results in avatars of higher realism than the prior art.
Second, the compositional nature supports editing of the individual components and 
enables the seamless transfer of features such as hairstyle or clothing between avatars.
These advancements open new possibilities for diverse applications, including virtual try-on.

\section{Related work}
\vspace{+1em}
\qheading{3D Avatar Creation From X.}
Creating realistic avatars is a long-standing challenge,
with many solutions for
building digital avatars from scans, videos, and images.  
Sophisticated capture systems are used to acquire high-quality 3D scans, which are turned into realistic, personalized, photorealistic avatars~\cite{borshukov2005realistic, alexander2010digital, alexander2013digital, seymour2017meet, lombardi2018deep, guo2019relightables}.
However, these methods are not scalable %
due to the high cost of building such systems and the sophisticated pipelines required for avatar creation. 
Consequently, there is great interest in creating avatars from easily captured images~\cite{tewari2017mofa,feng2021learning,zielonka2022towards,feng2018prn,PIXIE:3DV:2021} and monocular videos~\cite{grassal2022neural,zheng2022avatar,cao2022authentic,gao2022reconstructing,Feng2023DELTA}.
Such methods estimate 3D faces with the assistance of parametric models of the head~\cite{li2017learning,blanz1999morphable,basel} or full body~\cite{Pavlakos2019_smplifyx, li2017learning}.
Recent methods also learn {\em generative} 3D head models using only images~\cite{Chan2021eg3d,An2023PanoHeadG3}, allowing the creation of novel avatars from random noise inputs.
Additionally, there exist methods~\cite{feng2022capturing, ranade2022ssdnerf,rosu2022neural,li2023megane,Feng2023DELTA} that adopt a compositional approach to avatar modeling, enabling manipulation and control. 
Unlike these approaches, \modelname, requires only natural language descriptions to control the shape, texture, and accessories of a virtual avatar, making the creation and editing of realistic personal avatars accessible to a broad audience. 

\qheading{Text-Guided 3D General Object Generation.}
Following the success of recent \twoD text-to-image models~\cite{ramesh2022hierarchical,nichol2021glide,rombach2022high}, the generation of \threeD content from text is gaining attention
\cite{poole2023dreamfusion,lin2022magic3d,metzer2022latent,jain2022zero}.
Since paired text and 3D training data is scarce,
recent text-to-3D methods generate \threeD content by leveraging large, pretrained \twoD text-to-image models.
These approaches learn a \threeD representation using losses on the projected image in multiple \twoD views,
where pretrained models serve as frozen critics.
As an example, Contrastive Language--Image Pretraining (CLIP)~\cite{radford2021learning} is used to generate \threeD objects in the form of an occupancy network~\cite{sanghi2022clip}, a mesh~\cite{michel2022text2mesh,chen2022tango,mohammad2022clip}, and a NeRF~\cite{jain2022zero,wang2022clip}.
Similarly, DreamFusion~\cite{poole2023dreamfusion} and subsequent work \cite{lin2022magic3d,metzer2022latent,chen2023fantasia3d,shao2023control4d,wang2023prolificdreamer, huang2023tech} adopts a text-to-image diffusion model as a guide to optimize 3D object generation, %
significantly improving visual quality.
Despite the rapid progress in generating general objects, these methods suffer from visual artifacts when generating avatars (such as creating a person with multiple faces).

\qheading{Text-Guided \threeD Avatar Generation.}
Large, pretrained \twoD text-to-image models are also used for avatar creation. 
AvatarCLIP~\cite{hong2022avatarclip} uses the CLIP model to generate coarse body shapes, which are parameterized by the Skinned Multi-Person Linear (SMPL) model~\cite{loper2015smpl}.
DreamAvatar~\cite{cao2023dreamavatar} generates a \threeD human avatar from a given text prompt and SMPL body shape, where detailed shape and texture are learned under the guidance of a text-conditioned diffusion model.
These methods focus on full-body \threeD avatar generation, resulting in low-resolution faces with limited expression. 
T2P~\cite{zhao2023zero} uses CLIP supervision to optimize discrete attributes in a video game character creation engine.
As their method is confined to what can be represented by the game engine, their avatars are limited by the expressiveness of the artist-designed hairstyles and facial features. %
ClipFace~\cite{aneja2022clipface} enables text-guided editing of a textured \threeD morphable face model~\cite{li2017learning}, including expression and texture.
Describe3D~\cite{wu2023describe3d} synthesizes a \threeD face mesh with texture from a text description.
Neither approach explicitly models hair, and Describe3D resorts to manual post-processing to add hair.
To address this problem and boost the visual quality, DreamFace~\cite{zhang2023dreamface} employs a dataset of textures and artist-designed hairstyles, utilizing a CLIP model for component selection and a diffusion model for texture generation.
While DreamFace achieves realistic head avatars, it is limited to selecting pre-existing hairstyles from a manually curated gallery of artist-designed assets. 
Such an approach is expensive and does not scale.
Rodin~\cite{wang2022rodin} is a generative 3D face model trained from 100K synthetic 3D avatars.
They exploit CLIP's text and image embedding to enable text-guided generation and editing.
However, their results inherit the limited realism of the synthetic training data, and the approach does not disentangle the hair from the rest of the face.
Thus, it is difficult to change hairstyles without unwanted changes in facial appearance.
In contrast, our method generates realistic facial avatars using diffusion models without relying on artist-designed hairstyles.
Moreover, its compositional nature guarantees that edits to the hairstyle do not impact the face region.

\section{Method}
\begin{figure*}[t]
	\centering
	\includegraphics[width=\textwidth, trim={0cm, 0cm, 0cm, 0cm}]{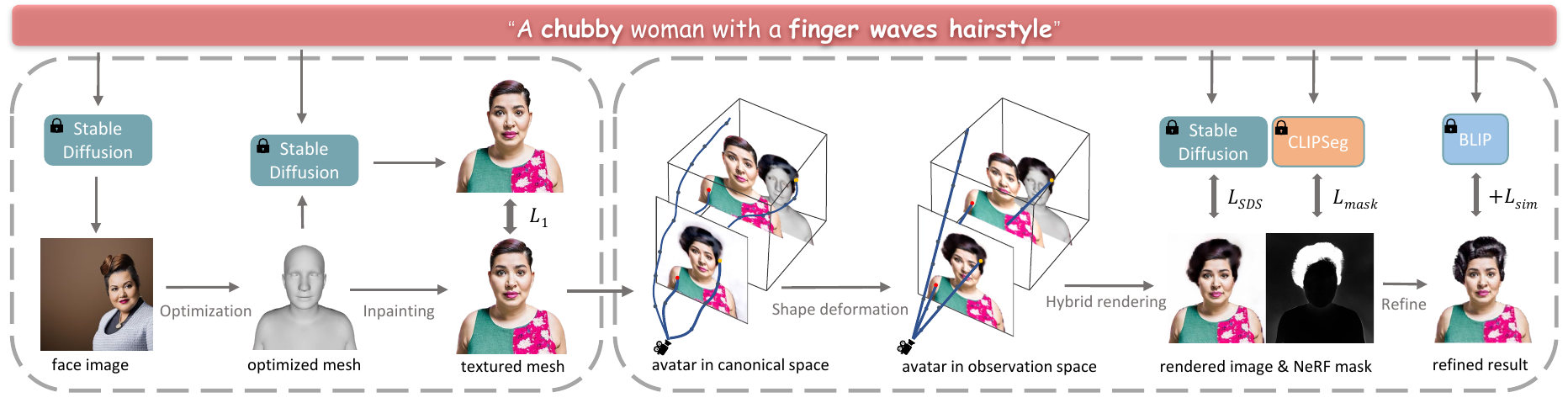}  
 \small{\hspace{1cm} Face generation (Sec.\ \ref{subsec:body}) \hspace{5cm} Hair generation (Sec.\ \ref{subsec:non-body})\hspace{2cm}}
 \vspace{-0.05in}
	\caption{Overview of \modelname. \modelname follows a sequential pipeline to generate realistic avatars. First, the text input is passed to Stable Diffusion to generate a single face image, which serves as a reference to obtain the geometry by SMPL-X fitting. We then adopt a texture painting approach inspired by \cite{richardson2023texture}, where the mesh is iteratively painted with a texture corresponding to the text using Stable Diffusion. Subsequently, style components such as the hair are modeled using NeRF in a latent space, with optimization guided by an SDS loss ($L_{\text{SDS}}$) and a mask loss ($L_{\text{mask}}$) with CLIPSeg segmentation, and finally refined in pixel space using $L_{\text{SDS}}$, $L_{\text{mask}}$, and an additional BLIP-based loss ($L_{\text{sim}}$). This hybrid modeling approach results in high-quality and realistic avatars.}
	\label{fig:pipeline}
 \vspace{-1em}
\end{figure*}

\modelname generates \threeD facial avatars with realistic hair and style components from text descriptions.
The overview of the pipeline is illustrated in Fig.\ \ref{fig:pipeline}.
Given a text description of a person's physical appearance, we first generate a corresponding image using Stable Diffusion~\cite{rombach2022high}.
We extract the \threeD geometry by fitting a SMPL-X model to this image. 
To generate the face texture, we follow the iterative inpainting approach of TEXTure~\cite{richardson2023texture},
which generates images from different viewpoints and projects them onto the surface.
The texture is generated using diffusion \cite{rombach2022high}, taking  the text, already-generated texture, and shape into account.
Conditioned on the generated face, we generate other  components such as hairstyles or clothing using NeRF, which we optimize using SDS constrained with a semantic mask generated with CLIPSeg.
The final refinement is done using a combination of SDS and BLIP-based losses.

\subsection{Preliminaries}
\label{subsec:preliminaries}
\vspace{+1em} 
\qheading{Parametric Body Model.}
\label{sec:smplx-canonical}
To model a realistic human avatar including the face and shoulders, we use the SMPL-X model~\cite{Pavlakos2019_smplifyx}.
SMPL-X is a parametric mesh model with identity $\shapecoeff \in \mathbb{R}^{\shapedim}$, pose $\posecoeff \in \mathbb{R}^{\posedim}$, and expression $\expcoeff \in \mathbb{R}^{\expdim}$ parameters that control the body and facial shape of the avatar.
SMPL-X provides a consistent, predefined topology with a set of vertices $\verts \in \mathbb{R}^{\numverts \times 3}$, which are computed by:
\begin{equation}
  \small
  \begin{aligned}
    \verts = & \  \vsmplx(\shapecoeff, \posecoeff, \expcoeff) \\
    = & \  \lbs(\deformfunc(\shapecoeff, \posecoeff, \expcoeff), \jointregressor(\shapecoeff), \posecoeff; \blendweights)
    \label{eq:smplx}
  \end{aligned}
\end{equation}
where $\lbs$ is a linear blend skinning function and $\blendweights \in \mathbb{R}^{\blendweightsdim}$ are the blend weights.
$\deformfunc$ represents the template mesh in a neutral, canonical pose:
\begin{equation}
   \small
   \deformfunc(\shapecoeff, \posecoeff, \expcoeff) = \template + \blendfunc(\shapecoeff, \posecoeff, \expcoeff),
    \label{eq:smplx_template}
\end{equation}
where $\blendfunc(\shapecoeff, \posecoeff, \expcoeff): \mathbb{R}^{\shapedim} \times \mathbb{R}^{\posedim} \times \mathbb{R}^{\expdim} \rightarrow \mathbb{R}^{\numverts \times 3}$ gives deformations of the template based on the shape, pose, and expression parameters.
$\jointregressor(\shapecoeff): \mathbb{R}^{\shapedim} \rightarrow \mathbb{R}^{\numjoints \times 3}$ is a joint regressor that takes in the identity shape parameters and produces the positions of the body joints.
SMPL-X builds a connection between the mean body shape $\template$ and a specific body shape $\verts$ with identity, pose, and expression information.
This can be formulated as a vertex-wise mapping from $\template$ to $\verts$.
Specifically, given $\bm{t}_i$ and $\bm{v}_i$ (the $i$-th row from $\template$ and $\verts$, respectively), the mapping is:
\begin{equation}
  \small
  \begin{aligned}
    \begin{bmatrix}
      \bm{v}_i^T \\
      1
    \end{bmatrix} & = 
    M_i(\shapecoeff, \posecoeff, \expcoeff) 
    \begin{bmatrix}
      \bm{t}_i^T \\
      1
    \end{bmatrix}.
  \end{aligned}
\end{equation}
Here, the function $M_i(\cdot)$ produces a $4\times 4$ matrix, $M_i(\shapecoeff, \posecoeff, \expcoeff) =$
\begin{equation}
\small
   \left(\sum_{k=1}^{\numjoints} \blendweight_{k,i} G_k(\posecoeff, \jointregressor(\shapecoeff))\right) 
    \begin{bmatrix}
      \bm{E} \!\! & \!\! B_i(\shapecoeff, \posecoeff, \expcoeff)^T \\
      \bm{0} \!\! & \!\! 1
    \end{bmatrix} ,
\end{equation} 
where $\blendweight_{k, i}$ is an entry from the blend skinning weights $\blendweights$ and $G_k(\posecoeff, \jointregressor(\shapecoeff))\in \mathbb{R}^{4\times 4}$ computes the world transformation for the $k$-th body joint.
$\bm{E}\in \mathbb{R}^{3\times 3}$ is the identity matrix, and  $\blendfunc_i$ computes the $i$-th row of the blend shapes.

\qheading{NeRF Representation.}
NeRF~\cite{mildenhall2020nerf} encodes a \threeD object as a continuous volumetric radiance field of color $\bm{c} \in \mathbb{R}^{|\bm{c}|}$ and density $\sigma \in \mathbb{R}$.
A NeRF is represented by a neural network $(\sigma, \bm{c}) = F_{\text{nerf}}(\bm{x}, \bm{p}; \bm{\Phi})$, where $\bm{x} \in \mathbb{R}^{3}$ is the location, $\bm{p} \in \mathbb{R}^{2}$ is the viewing direction, and $\bm{\Phi}$ are the learnable parameters of $F_{\text{nerf}}$.
Given a camera position, we estimate a \twoD image from the NeRF with volume rendering.
We denote a per-pixel ray $R(\ell) = \bm{o} + \ell\bm{d}$ by the origin $\bm{o} \in \mathbb{R}^{3}$, direction $\bm{d} \in \mathbb{R}^{3}$, and $\ell \in [\ell_n, \ell_f]$.
To discretize the rendering, we evenly split the rendering range into $n_{\ell}$ bins and randomly sample a $\ell_i$ for every bin with stratified sampling.
The volume rendering formulation for each pixel is:
\begin{equation}
  \small
  \begin{aligned}
    C(R(\ell)) &= \sum_{i=1}^{n_{\ell}} \alpha_i \bm{c}_i, \\ 
    \text{with} \ \ \alpha_i = \exp &\left(-\sum_{j=1}^{i-1} \sigma_j \Delta \ell_{j}\right)\left(1-\exp\left(-\sigma_i\Delta \ell_{i}\right)\right).
  \end{aligned}
  \label{nerf_rendering}
\end{equation}
Here, $\Delta \ell_i = \ell_{i+1} - \ell_i$ is the adjacent samples distance.

\qheading{Score Distillation Sampling.}
\label{sec:sds}
DreamFusion~\cite{poole2023dreamfusion} proposes Score Distillation Sampling (SDS) to guide \threeD content generation using pre-trained \twoD text-to-image diffusion models.
Following \cite{poole2023dreamfusion}, we denote the learned denoising function in the diffusion model as $\epsilon(\bm{Q}_t, \bm{y}, t)$.
Here, $\bm{Q}_t$ is the noisy image at timestep $t$.
SDS adopts the denoising function as a critic to update the \twoD rendering $\bm{Q}$ of the generated \threeD object across different viewpoints. 
The gradient is computed as:
\begin{equation}
   \small
   \nabla_{\bm{Q}} L_{\text{sds}}(\bm{Q}) = \mathbb{E}_{t, \bm{\epsilon}}\left[\bm{u}_t\cdot \left(\epsilon \left(\bm{Q}_t, \bm{y}, t\right) - \bm{\epsilon}\right) \right],
\end{equation}
where $\bm{u}_t$ is a weight at timestep $t$~\cite{ho2020denoising}.

\subsection{3D Face Generation}
\label{subsec:body}
To generate a 3D facial shape from a text description, we use a pre-trained Stable Diffusion model~\cite{rombach2022high} to synthesize a \twoD face image that semantically matches the given text. 
The descriptor keywords might include overweight, slim, muscular, or old.
Given the generated face image, we use an off-the-shelf landmark detector~\cite{bulat2017far} to obtain a set of facial landmarks $\{\bm{e}_i | \bm{e}_i\in \mathbb{R}^3\}_{i=1}^{n_e}$, where $n_e$ is the number of landmarks. We optimize the SMPL-X parameters using:
\begin{equation}
   \small
   (\shapecoeff^{*}, \posecoeff^{*}, \expcoeff^{*}) = \argmin_{\shapecoeff, \posecoeff, \expcoeff} \sum_{i=1}^{n_e}\norm{M_{\kappa(i)}(\shapecoeff, \posecoeff, \expcoeff)\bm{t}_{\kappa(i)} - \bm{e}_i}_{1}
\end{equation}
$\kappa(i)$ denotes the index of a vertex of the SMPL-X model that corresponds to the $i$-th landmark. 
Note that optimizing facial shape in this way results in full-body shape parameters, but the pose parameters are not well-constrained. Then, the final avatar shape is given by $\bm{V}^{*} = \vsmplx(\shapecoeff^{*}, \posecoeff^{c}, \bm{0})$, where $\posecoeff^{c}$ represents the body pose parameters corresponding to an ``A-pose''.

To generate a complete texture of the reconstructed face, we follow the iterative inpainting procedure of TEXTure~\cite{richardson2023texture}.
In each step, we generate an image $\bm{I}_i$ in the viewpoint of $\bm{p}_i$ using Stable Diffusion~\cite{rombach2022high} and project $\bm{I}_i$ back to the mesh surface according to the geometry $\bm{V}^{*}$ and $\bm{p}_i$.
The iterative inpainting can be denoted $\{\bm{A}_i\}_{i=0}^{n_p}$, where $\bm{A}_0$, $\{\bm{A}_i\}_{i=1}^{n_p-1}$, and $\bm{A}_{n_p}$ are the initial, intermediate, and final texture UV maps, respectively.
Denoting the differentiable mesh renderer as $R_m(\bm{A}, \bm{p}, \bm{V}^{*})$, the painting process can be summarized as:
\begin{equation}
   \small
   \bm{A}_{i} = \mathrm{argmin}_{\bm{A}} \norm{R_m(\bm{A}, \bm{p}_i, \bm{V}^{*})-\bm{I}_i}_{1}.
\label{inpaint}
\end{equation} %
To reduce cross-view conflicts in texture, we follow \cite{richardson2023texture} to take both the mesh geometry $\bm{V}^{*}$ and previous texture $\bm{A}_{i-1}$ information into account when generating the image $\bm{I}_i$.
This is achieved by iteratively applying depth-aware and inpainting diffusion models in the denoising process.
We also make the assumption that the face of the individual is approximately bilaterally symmetric and add an additional symmetry regularization term
\tdv{$L_{\text{sym}}$}. 
\tdv{
This term enforces similarity between the frontal face image and its horizontally flipped counterpart. Further information on this regularization and our texture generation process can be found in the \supmat
}

\begin{figure*}[t]
	\centering
	\includegraphics[width=\textwidth, trim={0cm, 0cm, 0cm, 0cm}]{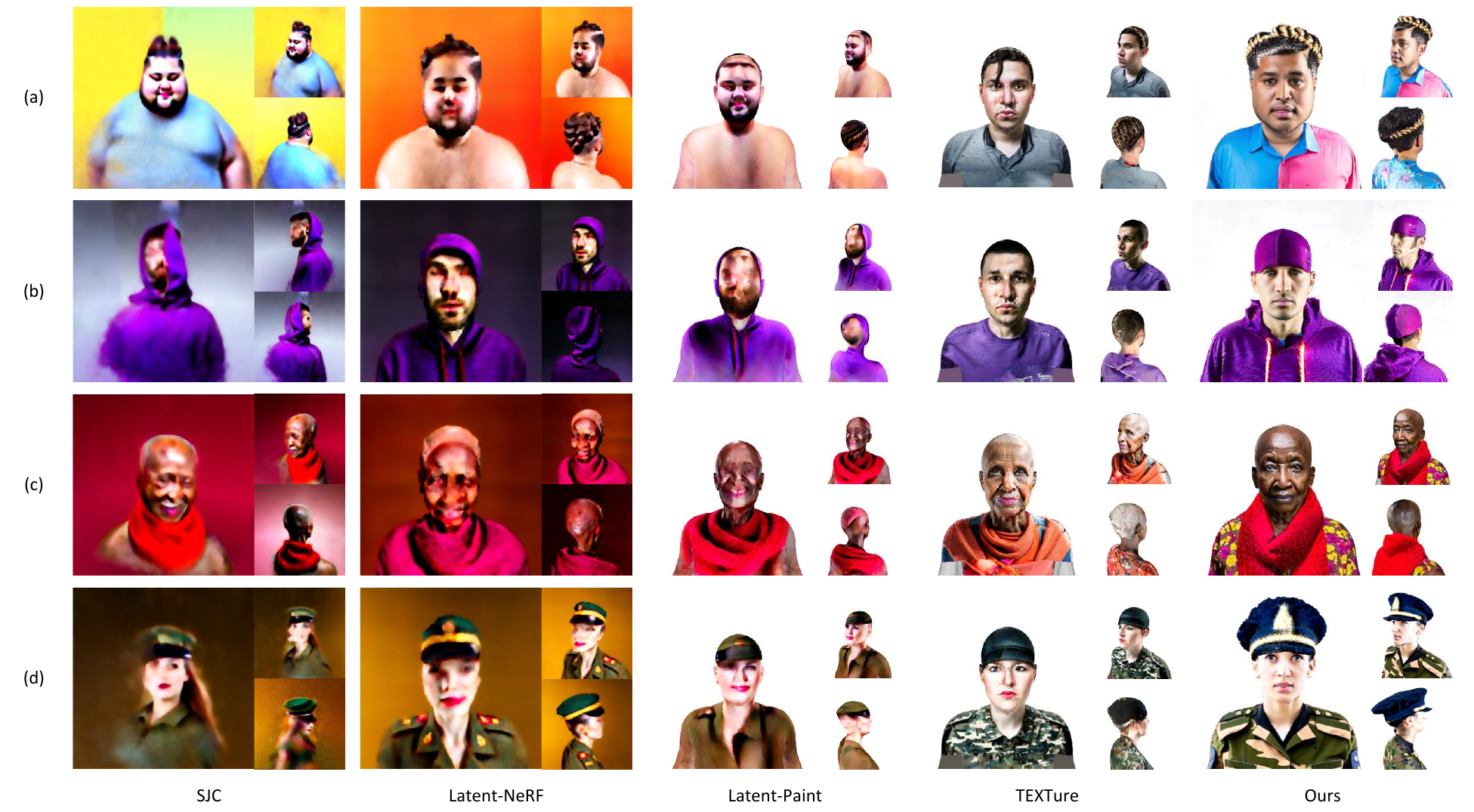}
    \vspace{-2em}
    \caption{
    Qualitative comparison with SOTA methods. 
    Text prompts: (a) ``An overweight man with a crown-braid hairstyle,'' (b) ``A man in a purple hoodie,'' (c) ``An old bald African woman wearing a wool scarf,'' (d) ``A young woman in a military hat.'' }
    \vspace{-1em}
\label{fig:comparison}
\end{figure*}
\subsection{Hair, Clothing, and Accessory Generation}
\label{subsec:non-body}
\vspace{+1em}
\qheading{Canonicalization.}
\tdv{
Building upon the generated face, represented as a textured mesh, we learn a separate NeRF model for attributes like hair, clothing, or accessories. The NeRF model is built in a canonical space, which is constructed around the SMPL-X template mesh $\template$, enabling the animation and transfer of the non-face parts.
}
For body mesh $\bm{V}$ and its corresponding parameters $\shapecoeff$, $\posecoeff$, and $\expcoeff$, we follow previous work~\cite{chen2021animatable, feng2022capturing, cao2023dreamavatar} to map the points from observation space to canonical space ($\bm{x} \rightarrow \bm{x}^{c}$):
\begin{equation}
   \small
   \bm{x}^{c} = \sum_{\bm{v}_i \in \mathcal{N}\left(\bm{x}\right)} \frac{\omega_i(\bm{x}, \bm{v}_i)}{\omega(\bm{x}, \bm{v}_i)} M_{i}(\bm{0}, \posecoeff^{c}, \bm{0}) \left(M_{i}(\shapecoeff, \posecoeff, \expcoeff)\right)^{-1}\bm{x},
\end{equation}
where $\mathcal{N}(\bm{x})$ is the set of nearest neighbors of $\bm{x}$ in $\bm{V}$.
The weights are computed as:
\begin{equation}
   \small
   \begin{aligned}
      \omega_i(\bm{x}, \bm{v}_i) &=\exp \left(-\frac{\left\|\bm{x}-\bm{v}_{i}\right\|_2 \left\|\bm{\blendweight}_{\xi(\bm{x})}-\bm{\blendweight}_{i}\right\|_2}{2 \tau^{2}}\right), \text{ and} \\
      \omega(\bm{x}, \bm{v}_i) &=\sum_{\bm{v}_i \in \mathcal{N}(\bm{x})} \omega_i(\bm{x}),
   \end{aligned}
\end{equation}
where $\xi(\bm{x})$ is the index of the vertex in $\bm{V}$ that is closest to $\bm{x}$. $\bm{w}_i$ is the $i$-th column of $\bm{W}$, and $\tau$ is 0.2.

\qheading{Mesh-Integrated Volumetric Rendering.}
Conditioned on the textured face mesh, we learn NeRF models with mesh-integrated volume rendering following \cite{feng2022capturing}.
Specifically, when a ray $R(\ell)$ is emitted from the camera center $\bm{o}$ and intersects the mesh surface, we set $\ell_f$ such that $R(\ell_f)$ represents the first intersection point.
The texture color at $R(\ell_f)$, denoted by $\bm{c}^{*}$, is then used in the volume rendering by extending Eqn.\ \ref{nerf_rendering}:
\begin{equation}
  \small
  \begin{aligned}
    C(R(\ell)) &= \left(1-\sum_{i=1}^{n_{\ell}-1} \alpha_i\right) \bm{c}^{*} + \sum_{i=1}^{n_{\ell}-1} \alpha_i \bm{c}_i, \\ 
    \text{with} \ \ \alpha_i = \exp &\left(-\sum_{j=1}^{i-1} \sigma_j \Delta \ell_{j}\right)\left(1-\exp(-\sigma_i\Delta \ell_{i})\right).
  \end{aligned}
\end{equation}
For a ray that does not intersect the mesh surface, the computation of aggregated color follows Eqn.\ \ref{nerf_rendering}. 
Unlike in a traditional NeRF representation, we model NeRF in a latent space to accelerate the learning process. Specifically, $\bm{c}$ and $\bm{c}^{*}$ represent 4-dimensional latent features.
While the features $\bm{c}_i$ are optimized, the latent feature on the mesh surface $\bm{c}^{*}$ is obtained by running the stable diffusion encoder~\cite{rombach2022high} on the RGB rendering of the textured mesh.
After volumetric integration, the resulting feature image is decoded with the Stable Diffusion model. 

To train a NeRF model for a specific style component, we rely on CLIPSeg~\cite{lueddecke22_cvpr} for spatial guidance. 
Taking hair as an example, we use CLIPSeg with a keyword of ``hair'' to segment the hair region in the image generated by rendering the hybrid mesh--NeRF model.
This hair segmentation mask $\bm{\Omega}$ indicates the hair and non-hair regions.
We use $\bm{\Omega}$ to guide the NeRF model to focus on the representation of objects within the masked region while discouraging it from learning geometry outside the region.
This is useful to prevent the NeRF from modeling the face, when it should only represent hair.
We use the following mask-based loss:
\begin{equation}
   \small
   L_{\text{mask}} = \norm{\bm{\Omega} - \hat{\bm{\Omega}}}_{1} ,
\end{equation}
where $\hat{\bm{\Omega}}$ is the rendered NeRF mask, obtained by sampling rays for all pixels of the entire image.
The computation of a mask value at pixel location $R(\ell)$ is given by:
\begin{equation}
   \small
   \Omega_i\left(R(\ell), n_{\ell}-1\right) = \sum_{i=1}^{n_{\ell}-1} \alpha_i.
\end{equation}

To prevent floating ``radiance clouds,'' we incorporate the sparsity loss from \cite{metzer2022latent} into the training of the NeRF:
\begin{equation}
  \small
  \begin{aligned}  
     L_{\text{sparse}} = \sum_{i} F_{\text{BE}} \left(\Omega_i\left(R(\ell), n_{\ell}\right)\right),
  \end{aligned}
\end{equation}
where $F_{\text{BE}}(a) = -a \ln a - (1-a) \ln (1-a)$ is a binary entropy function.
The total training loss function for the latent NeRF models is:
\begin{equation}
  \small
  \begin{aligned}  
     L_{\text{NeRF}} = L_{\text{sds}} + \lambda_{\text{mask}}L_{\text{mask}} + \lambda_{\text{sparse}}L_{\text{sparse}} .
  \end{aligned}
\end{equation}

\qheading{Style Component Refinement in RGB Space.} 
While latent NeRF models can be used to learn reasonable geometry and texture, we have observed that adding further refinement in RGB space using a combination of the SDS loss and a loss based on BLIP~\cite{li2022blip} improves local detail.
Our BLIP loss, $L_{\text{sim}}$, measures the similarity of high-level visual and text features. 
Maximizing their similarity encourages the NeRF model to capture additional details, including structure, texture, and semantic content from the text description, leading to visually appealing results, see Fig.\ \ref{fig:ablation_refine}.

To perform the refinement, we append an additional linear layer to the NeRF model that converts the 4D latent feature into a 3D color representation~\cite{KevinTurner2022}.
The initial weights of this layer are computed using pairs of RGB images and their corresponding latent codes over a collection of natural images.
Let $\bm{z}_{\text{img}}$ and $\bm{z}_{\text{text}}$ be the embeddings of the rendered image and text prompt, then the similarity loss is:
\begin{equation}
  \small
  \begin{aligned}  
     L_{\text{sim}} = -\frac{\bm{z}_{\text{img}}^T\bm{z}_{\text{text}}}{\|\bm{z}_{\text{img}}\|\cdot \|\bm{z}_{\text{text}}\|},
  \end{aligned}
\end{equation} 
and the learning objective in refinement stage is:
\begin{equation}
  \small
  \begin{aligned}  
     L_{\text{refine}} = L_{\text{NeRF}} + \lambda_{\text{sim}}L_{\text{sim}} .
  \end{aligned}
\end{equation}
More implementation details are included in \supmat

\section{Experiments}

\label{sec:experiments}
We evaluate \modelname through 
\tdv{
1) comparisons with state-of-the-art (SOTA) methods for text-guided generation, 2) an online perceptual study, 3) quantitative evaluation,  4) the application of try-on and animation of generated avatars, and 5) ablation studies exploring our design choices.
}
To evaluate \modelname's ability to generate diverse compositional avatars, we need text prompts that are also compositional.
These text prompts may include facial attributes (\emph{e.g.}~overweight, slim, muscular), hairstyles (\emph{e.g.}~bun, afro, braid), clothing (\emph{e.g.}~jacket, wool scarf, hat), and more.
The candidate attributes and styles words are taken from a dataset of faces~\cite{liu2015deep}, a dataset of hairstyles~\cite{wei2022hairclip}, and other online 
sources\footnote{\href{https://7esl.com/vocabulary-clothing-clothes-accessories/}{\tt vocabulary-clothing-clothes-accessories}}\footnote{\href{https://7esl.com/types-of-hats/}{\tt types-of-hats}}.
In total, these attributes produce up to 3,300 text-prompt combinations, ensuring rich diversity.

\subsection{Comparisons with SOTA Methods}
We compare \modelname with four SOTA methods.
Two are solely based on NeRF representations (SJC~\cite{wang2022score} and Latent-NeRF~\cite{metzer2022latent}) and two are based on mesh painting techniques (Latent-Paint~\cite{metzer2022latent} and TEXTure~\cite{richardson2023texture}).
Figure \ref{fig:comparison} shows generated examples obtained from four diverse text prompts describing various personal characteristics, hairstyles, and clothing.
Notably, all methods successfully generate avatars with recognizable features that semantically align with the text, such as gender, color, and clothing items.
However, SJC and Latent-NeRF produce visually distorted and incomplete avatars, primarily due to flawed geometry and low-resolution textures.
While Latent-Paint incorporates a mesh as a shape prior, leading to reasonable proportions, the textures still suffer from blurriness and a lack of cross-view consistency.
TEXTure demonstrates good texture quality but is limited by the mesh topology; it cannot non-body components like the crown-braid hairstyle and hoodie.
In contrast, \modelname generates more realistic and natural avatars with strong cross-view consistency.
Our text-generated avatars exhibit detailed appearances, including diverse hairstyles and accessories (see Sup. Mat.).%

\subsection{Perceptual Study}
\label{perceptual_study}
\begin{figure}[tbh]
\centerline{\includegraphics[width=\linewidth]{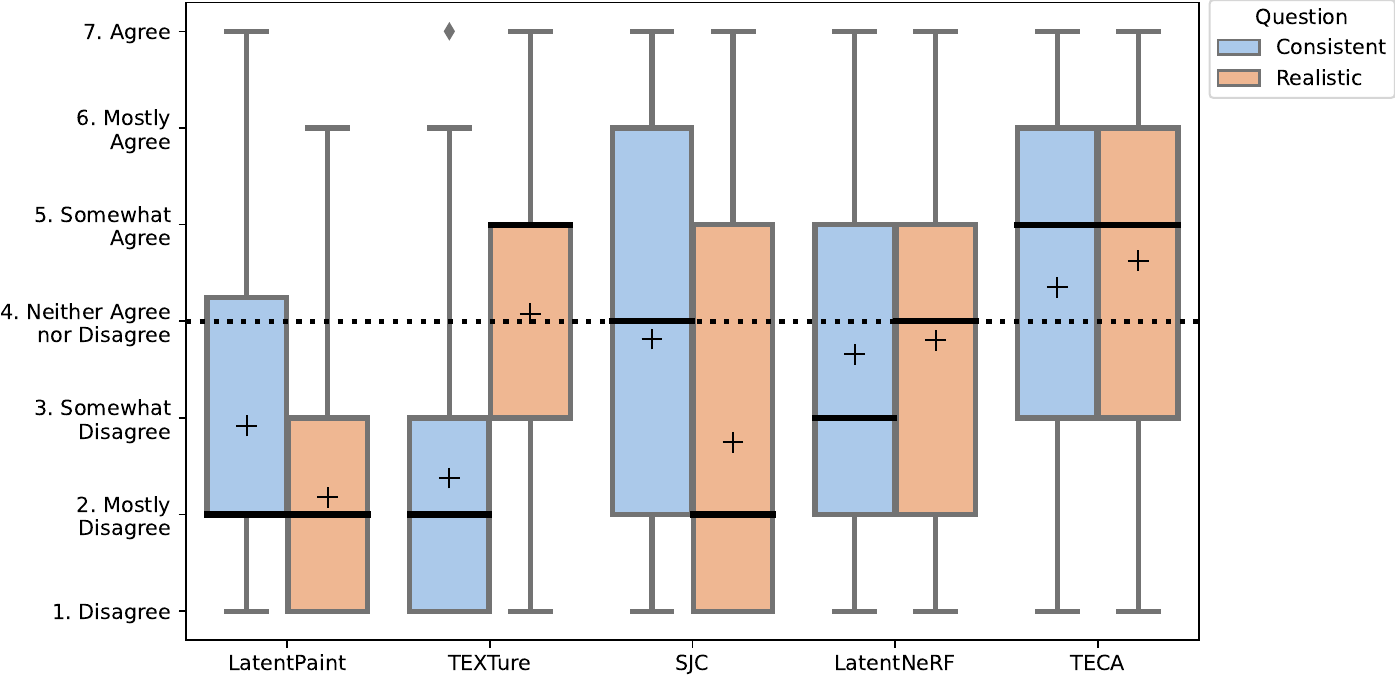} }
\caption{
\tdv{
A box-and-whisker plot of the perceptual study results. Users were asked the questions 
1) “The appearance of the avatar in the video matches the text description below it.” (blue color) and 
2) “The avatar in the video is visually realistic” (orange color),
`+' corresponds to the mean.}
}
	\vspace{-1.em}
	\label{fig:perceptual}
\end{figure}

We conducted an online perceptual study to rate avatars synthesized by \modelname and the baseline methods, SJC, LatentNeRF, LatentPaint, and TEXTure.
Participants on Amazon Mechanical Turk were presented with videos of the generated avatars and asked to rate the visual realism and consistency with the text prompt using a seven-point Likert scale.
We showed each participant the same thirty prompts but shuffled which method’s avatar the participants saw for each prompt.
Each participant was shown an equal number of avatars synthesized by each method.
A total of 150 responses were collected, out of which 52 (35\%) participants passed all of the catch trials and were included in the study. 
\tdv{
We applied the non-parametric Friedman test and then performed the Nemenyi post-hoc test to identify pairwise differences. The results are shown in Fig.\ \ref{fig:perceptual} and only our method receives, on average, positive ratings to both questions.
}
See \supmat 
for more details.

\subsection{Quantitative Evaluation}

\tdv{
We also quantitatively compare our method with other state-of-the-art methods. Specifically, we assess the semantic matching between text and avatar using the CLIP~\cite{radford2021learning} score and evaluate the generated avatar quality through the Fréchet Inception Distance (FID)~\cite{Seitzer2020FID}. 
To compute the CLIP score, we convert the videos used in Section \ref{perceptual_study} into images and employed CLIP~\cite{radford2021learning} to compute the cosine distance between these images and their respective text prompts.  The results are shown in Table \ref{tab:quantitative_evaluation}.  
Our method achieves the highest semantic consistency, indicating its ability to accurately represent text descriptions. 
For FID, we generated 200 avatars using 40 different text prompts, with 5 random seeds for each prompt. Each avatar was rendered from 50 different views, resulting in a total of 10,000 images for evaluation. The ground truth distribution was based on the first 10,000 images from the Flickr-Faces-HQ Dataset (FFHQ) ~\cite{karras2019style}.
The FID scores are shown in Table \ref{tab:quantitative_evaluation}. TECA has the lowest FID score, indicating its superior image quality relative to the other methods. 
} 

\begin{table}[ht]
\small
\centering
\begin{tabular}{ccc}
\toprule
Method & CLIP Score $\uparrow$ & FID$\downarrow$\\
\midrule
SJC \cite{wang2022score} & 0.2979 & 27.50 \\
Latent-NeRF \cite{metzer2022latent} & 0.3025 & 24.49 \\
Latent-Paint \cite{metzer2022latent} & 0.2854 & 38.33 \\
TEXTure \cite{richardson2023texture} & 0.2761 & 29.51 \\
Ours & \textbf{0.3213} &  \textbf{14.98} \\
\bottomrule
\end{tabular}
\caption{Quantitative evaluation results. Higher CLIP score indicates better consistency between the text prompts and generated avatars, lower FID indicates higher realism of the avatars.}
\vspace{-1.em}
\label{tab:quantitative_evaluation}
\end{table}

\subsection{Applications: Try-on and Animation}
\begin{figure}[t]
	\centering
	\includegraphics[width=\linewidth]{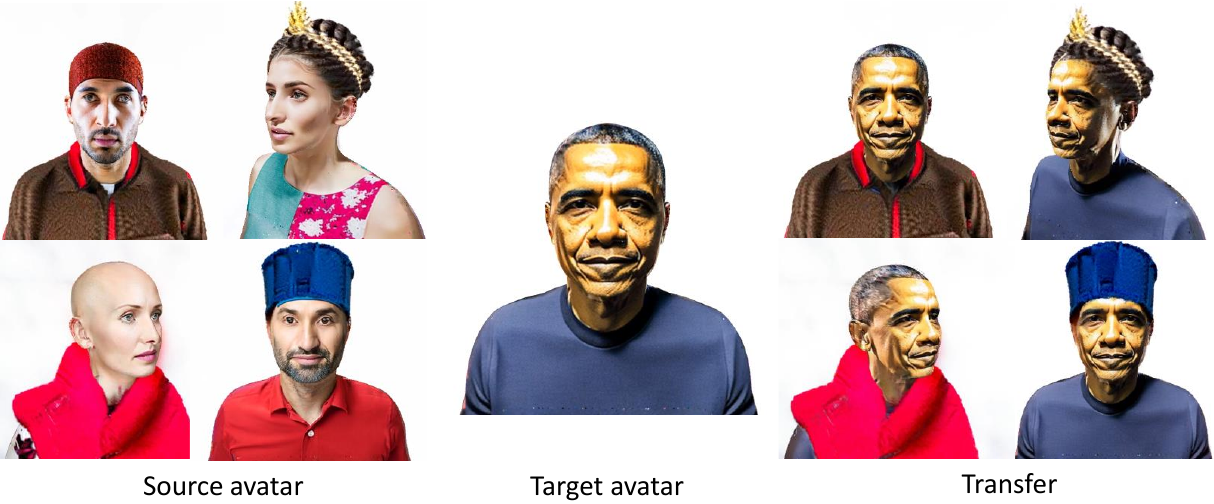} \\  
	\caption{\modelname Try-On.
    The hairstyle, hat, scarf, or clothing is transferred from the source avatars to the target avatar.
    }
	\vspace{-0.7em}
	\label{fig:application}
\end{figure}

Since \modelname is compositional, components like hairstyles and accessories can be transferred between avatars.
As shown in Fig.\ \ref{fig:application}, \modelname can transfer a brown pullover, crown-braid hairstyle, red scarf, or blue hat to the target avatar.
The non-face components adjust in size to adapt to a new head shape.
This makes our generated hairstyles and accessories highly versatile.
For instance, users can input their own avatars and transfer a learned hairstyle to it.

\begin{figure}[t]
	\centering
	\includegraphics[width=0.88\linewidth]{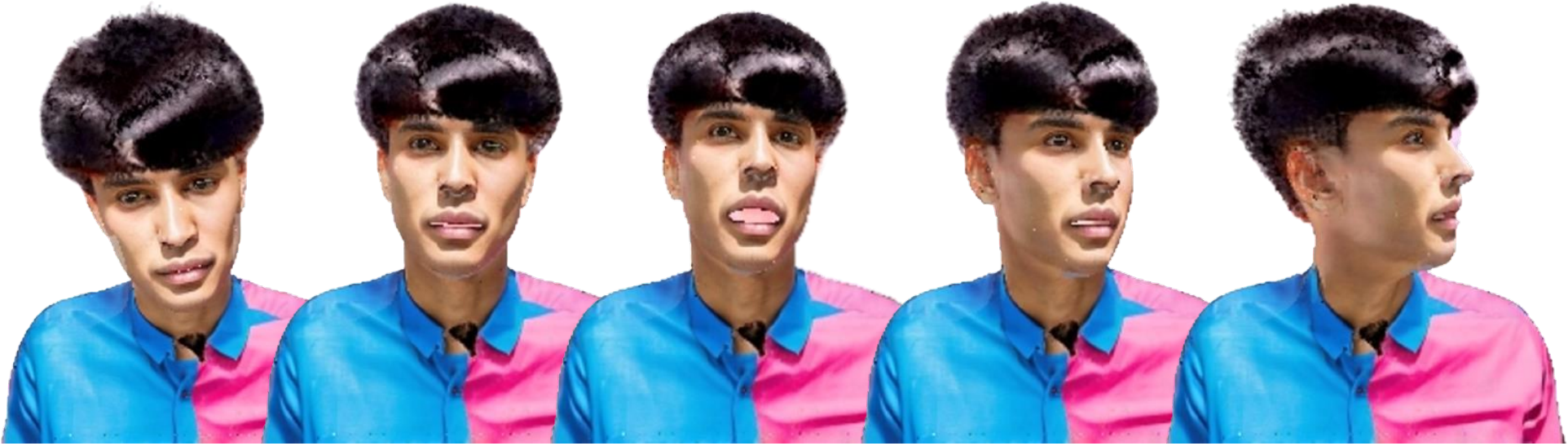} \\  
        \vspace{-0.5em}
	\caption{\modelname Animation.
    The generated avatar is animated to different poses and expressions.
    }
	\label{fig:animation}
\end{figure}

\tdv{
Leveraging the SMPL-X model, we also gain the flexibility to animate the avatar across various poses and expressions. As detailed in Sec.~\ref{subsec:non-body}, the NeRF component has the capacity to synchronize its movement with the dynamics of the  SMPL-X body. 
Displayed in Fig.~\ref{fig:animation}, our generated avatar can be animated with varying head poses and expressions. Notice the transition from a neutral to an open-mouth expression (middle image), where the interior mouth region is absent. 
This limitation could be addressed by inpainting the inside mouth region using diffusion models. Further instances showcasing this including face editing results are available in the \supmat
}

\subsection{Ablation Experiments}
\vspace{1em}

\begin{figure}[t]
	\centering
    \includegraphics[width=\linewidth]{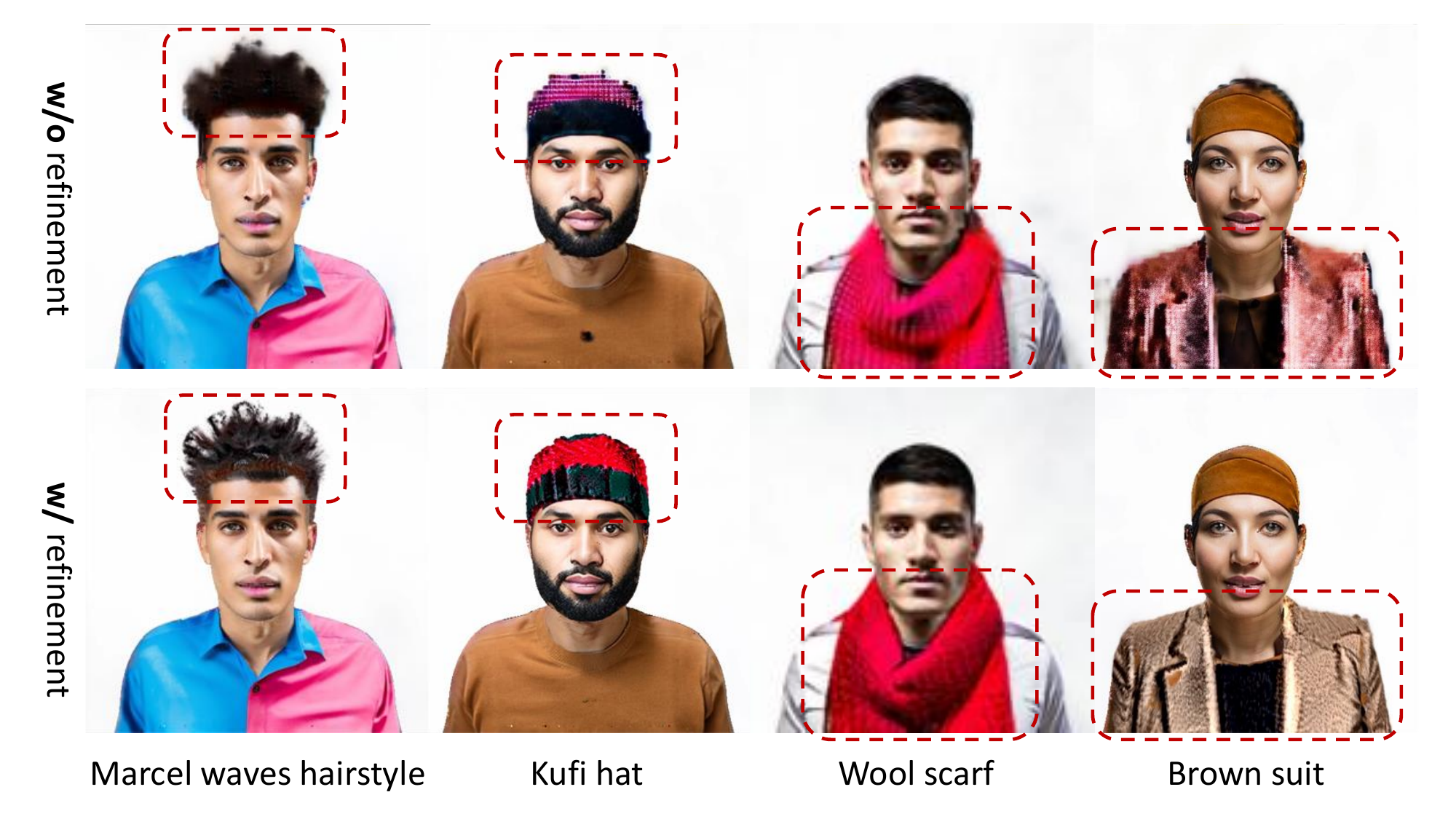} \\  
        \vspace{-0.5em}
	\caption{
 Comparison of results between unrefined (top) and refined (bottom) style components. 
 The refinement improves the details of hair, hat, scarf, and clothing. The refined components are indicated by red dotted line boxes.}
	\label{fig:ablation_refine}
\end{figure}

\qheading{Non-Face Refinement.} 
We investigate the effect of refining non-face details using a combination of SDS  and BLIP losses.
Figure \ref{fig:ablation_refine} illustrates the difference between refined and non-refined style components.
The results demonstrate that the refinement produces more detail, noticeably enhancing the overall visual quality.

\begin{figure}[t]
	\centering
	\includegraphics[width=\linewidth]{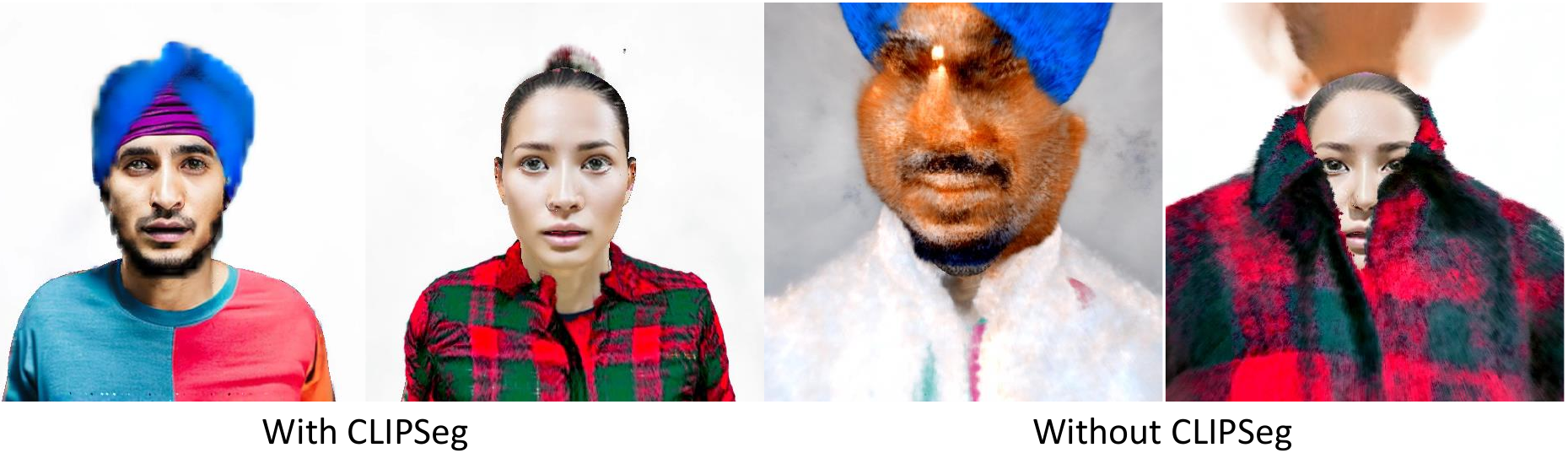} \\  
        \vspace{-0.5em}
	\caption{
 Ablation of CLIPSeg. Without the segmentation information, NeRF learns the entire avatar instead of the components.
 }
	\label{fig:ablation_clipseg}
\end{figure}

\qheading{CLIPSeg Segmentation.}
The segmentation loss prevents NeRF from trying to represent the entire avatar and focuses it on representing a specific part.
Without the loss, the results are significantly worse; see Fig.\ \ref{fig:ablation_clipseg}. 

\newpage
\section{Discussion and Limitations} 
\vspace{1em}

\qheading{Segmentation with CLIPSeg.} 
\begin{figure}[t]
	\centering
	\includegraphics[width=\linewidth]{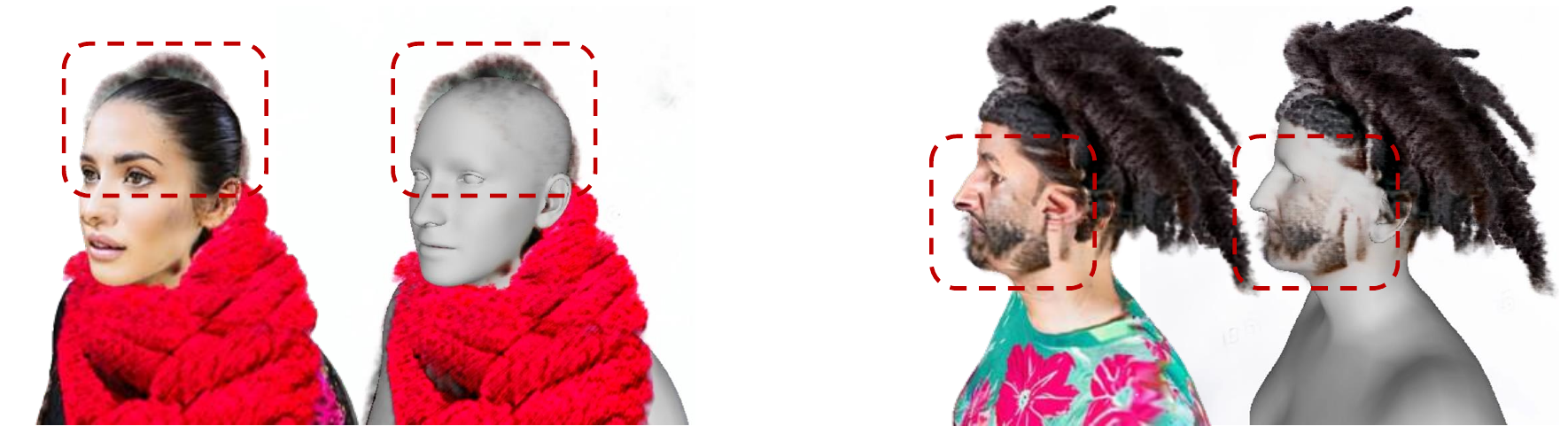} \\  
	\vspace{-0.5em}
	\caption{Failure cases showing the impact of poor segmentation. 
 } 
	\label{fig:limitation_seg}
\end{figure}
To guide the learning process, we use CLIPSeg to obtain masks for the region of interest, such as hair or clothing.
This encourages the NeRF to focus on learning specific components rather than on the entire avatar.
Our method's effectiveness is contingent on the segmentation quality.
If CLIPSeg encounters challenges, flawed NeRF representations may result, such as floating points in the face region, as shown in Fig.\ \ref{fig:limitation_seg}.

\qheading{Performance of Diffusion Models.}
Our results are constrained by the capabilities and biases of pretrained diffusion models because they provide the semantic information for avatar generation.
\tdv{
\qheading{Dynamics.}
TECA's ability to animate avatars via the SMPL-X parametric model highlights its potential, yet addressing complex dynamics in elements like hair and clothing calls for further exploration.
}
\qheading{Relighting.}
Our model does not support relighting in new environments, as the learned RGB color for the face texture and NeRF-based hair or accessories is baked with lighting.
Further work is needed to disentangle albedo and lighting attributes to enable relighting.

\section{Conclusion} 
We presented \modelname, an innovative method for generating realistic \threeD facial avatars with hair and accessories from text descriptions. 
By adopting a compositional model and using distinct representations for different components, we addressed the limitations of existing methods in terms of realism, shape fidelity, and capabilities for editing.
Our experimental results demonstrate the superior performance of \modelname compared to state of the art, delivering highly detailed and editable avatars.
Further, we demonstrated the transfer of hairstyles and accessories between avatars.


\qheading{Disclosure} This work was partially supported by the Max Planck ETH Center for Learning Systems. 
MJB has received research gift funds from Adobe, Intel, Nvidia, Meta/Facebook, and Amazon.  MJB has financial interests in Amazon, Datagen Technologies, and Meshcapade GmbH.  While MJB is a consultant for Meshcapade, his research in this project was performed solely at, and funded solely by, the Max Planck Society.

\section{Appendix}
\subsection{Implementation Details}
\label{implementation_details}

\begin{figure}[t]
	\centering
	\includegraphics[width=0.9\linewidth]{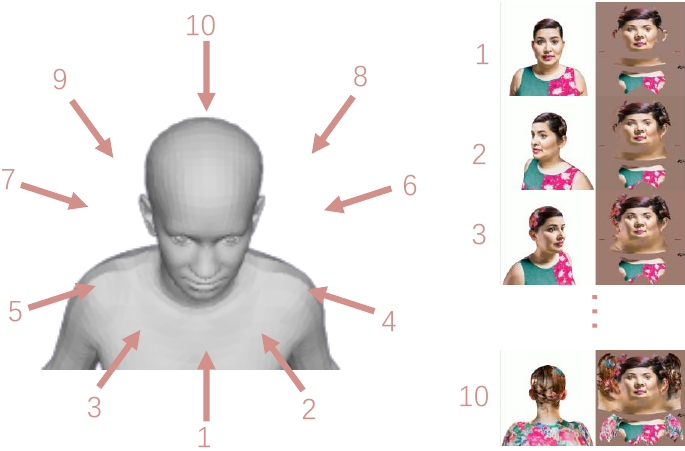} \\  
        \vspace{-0.5em}
	\caption{\tdv{Left: The ten virtual camera views used for the texture optimization process. Right: The rendered images (1st column) and the corresponding UV maps (2nd column).}}
	\vspace{-0.5em}
	\label{fig:muitiviews_texture}
\end{figure}

For the SMPL-X model~\cite{Pavlakos2019_smplifyx}, we use $\shapedim=300$ and $\expdim=100$, and we use 68 facial landmarks to fit the SMPL-X shape to the reference images, ($n_e=68$).
In the SMPL-X optimization process, we incorporate shape and expression regularization~\cite{Pavlakos2019_smplifyx} with a weight of $5e-5$.
For texture generation on the mesh, we use $n_p=10$ viewing directions. 
\tdv{
The mesh texture is optimized in an iterative manner from various view angles, following the sequence illustrated in Fig.\ \ref{fig:muitiviews_texture}.
The symmetry loss $L_{sym}$ is applied during this process, leveraging our assumption that the face is approximately symmetric. 
Specifically, for front and back views (No.\ 1 and No.\ 10), we apply an $L_2$ loss between the rendered image $R_m(\bm{A}, \bm{p}_i, \bm{V}^{*})$ that renders the geometry $\bm{V}^{*}$ in the view direction $\bm{p}_i$ based on the UV map $\bm{A}$ and the horizontally flipped version of the Stable Diffusion~\cite{rombach2022high} generated image $\bm{I}^{\prime}_i$. 
For right views (Nos.\ 3, 5, 7, and 9), we implement the $L_2$ loss by comparing the rendered images and the corresponding left views (Nos.\ 2, 4, 6, and 8). 
}
During NeRF training, we \tdv{sample 96 points each ray} ($n_{\ell}=96$). During the refinement stage, we increase the number of sampled points $n_{\ell}$ to 128.
In the optimization, we employ a latent code with dimensions of $64\times64\times4$. During refinement, we render the image at a resolution of $480\times480\times3$.
The network used in the NeRF training process comprises three fully connected layers, and the adapter from the latent space to the pixel space is a one-layer MLP following \cite{KevinTurner2022}.
Other hyperparameters include $\tau=0.1$, $|\mathcal{N}(\bm{x})|=6$, $\ell_{n}=-1$, and $\ell_{f}=1$. 

\tdv{
For the Stable Diffusion model~\cite{rombach2022high} and its depth-aware and inpainting variants, we use the implementations available on HuggingFace\footnote{\href{https://huggingface.co/stabilityai/stable-diffusion-2}{\tt stabilityai/stable-diffusion-2}}\footnote{\href{https://huggingface.co/stabilityai/stable-diffusion-2-depth}{\tt stabilityai/stable-diffusion-2-depth}}\footnote{\href{https://huggingface.co/stabilityai/stable-diffusion-2-inpainting}{\tt stabilityai/stable-diffusion-2-inpainting}}.
For the BLIP model~\cite{li2022blip}, we use the released model\footnote{\href{https://storage.googleapis.com/sfr-vision-language-research/BLIP/models/model_base.pth}{\tt BLIP/models/model\_base$.$pth}} from the LAVIS project~\cite{li-etal-2023-lavis}.
}
We employ the Adam optimizer~\cite{kingma2014adam} with a learning rate of $0.01$ for texture optimization, while for other optimizations, we use a learning rate of $0.001$.
For loss weights, we fix $\lambda_{\text{sym}}=0.5$, $\lambda_{\text{mask}}=0.1$, $\lambda_{\text{sparse}}=0.0005$, and $\lambda_{\text{sim}}=1$.
The average run time for avatar generation is currently around three hours on an A100 GPU.

\begin{figure*}[t]
	\centering
	\includegraphics[width=\textwidth, trim={0cm, 0cm, 0cm, 0cm}]{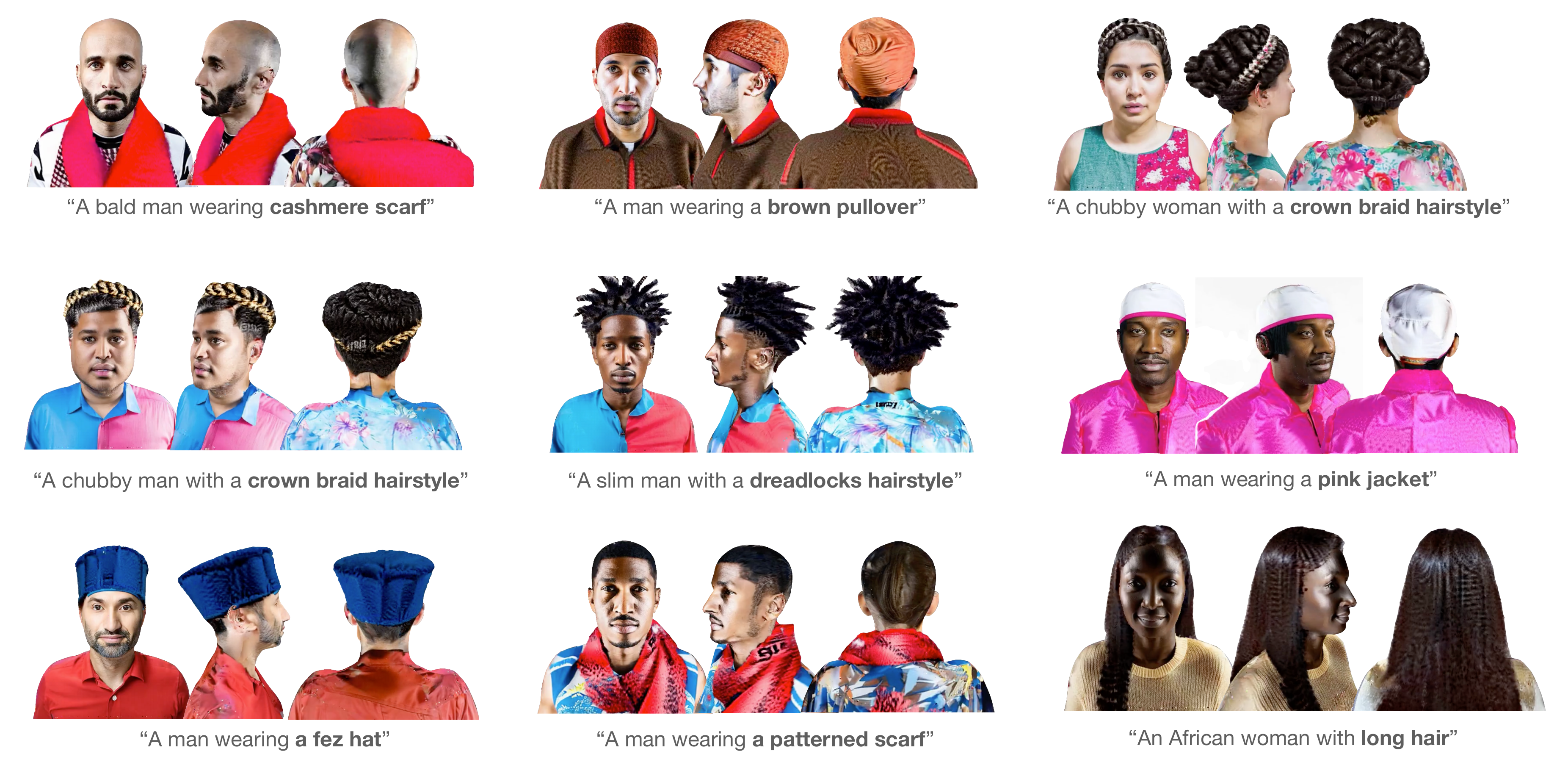}  
 \vspace{-1em}
	\caption{Additional examples for generated avatars by our method.}
	\label{fig:more}
\end{figure*}
\begin{figure*}[t]
	\centering
	\includegraphics[width=\textwidth, trim={0cm, 0cm, 0cm, 0cm}]{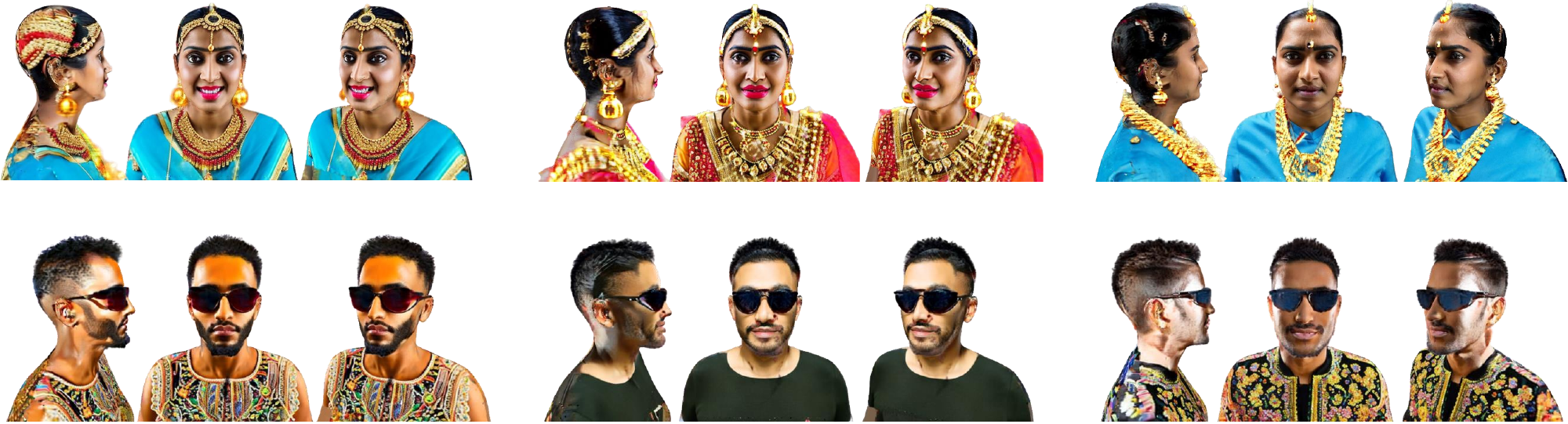}  
 \vspace{-1em}
	\caption{Additional examples of accessories such as earrings, necklaces, and glasses.}
	\label{fig:more_accessories}
 \vspace{-1em}
\end{figure*}
\subsection{Perceptual Study Details} \label{sec:perceptual_appendix}
In our survey, we used a random selection process to create a set of thirty unique prompt combinations, each consisting of five elements.
These elements were: 1) gender (male or female); 2) a color\footnote{\href{https://raw.githubusercontent.com/wty-ustc/HairCLIP/main/README.md}{\tt HairCLIP/main/README.md}}; 3) a hairstyle\footnote{\href{https://raw.githubusercontent.com/wty-ustc/HairCLIP/main/mapper/hairstyle_list.txt}{\tt HairCLIP/main/mapper/hairstyle\_list.txt}} or hat\footnote{\href{https://7esl.com/types-of-hats/}{\tt 7esl.com/types-of-hats}}, weighted equally; 4) another color; and 5) a type of upper-body clothing\footnote{\href{https://7esl.com/vocabulary-clothing-clothes-accessories/}{\tt 7esl.com/vocabulary-clothing-clothes-accessories}}.
These combinations were used to construct a prompt in the form of ``A [1] with a(n) [2] [3] wearing a(n) [4] [5]''.

To mitigate potential interaction effects resulting from participants' unfamiliarity with the styles presented, we included an image from the internet that represented the hairstyle or type of hat described in the prompt; the image was displayed next to the avatar video.
Participants were then asked to rate their agreement with two statements: 1) ``The avatar in the video is visually realistic'' and 2) ``The appearance of the avatar in the video matches the text description below it.''
To determine whether a participant successfully passed the catch trials, we examined their ratings for both questions.
Participants were considered to have passed if they rated both questions with greater-than-neutral agreement or greater-than-neutral disagreement on all five constant manually curated high-quality samples and catastrophic generation failures, respectively.
\tdv{
A total of 150 responses were collected, out of which 52 (35\%) participants passed all of the catch trials and were included in the study. The response distributions failed the Shapiro--Wilk normality test, so we applied the non-parametric Friedman test, which indicated that the method used to generate the avatar had a statistically significant effect on the outcomes of both study questions. Subsequently, we performed the Nemenyi post-hoc test to identify pairwise differences.
Using a significance level ($\alpha$) of $0.05$, the perceived {\em realism} of \modelname was determined to be significantly different than that of all baselines other than TEXTure, and the {\em text consistency} was determined to be significantly different than that of all baselines.
}
These findings confirm our initial expectations regarding the strengths of each method and support the value of our proposed combination of mesh-based and NeRF-based representations.

\begin{figure*}[t]
	\centering
	\includegraphics[width=\textwidth, trim={0cm, 0cm, 0cm, 0cm}]{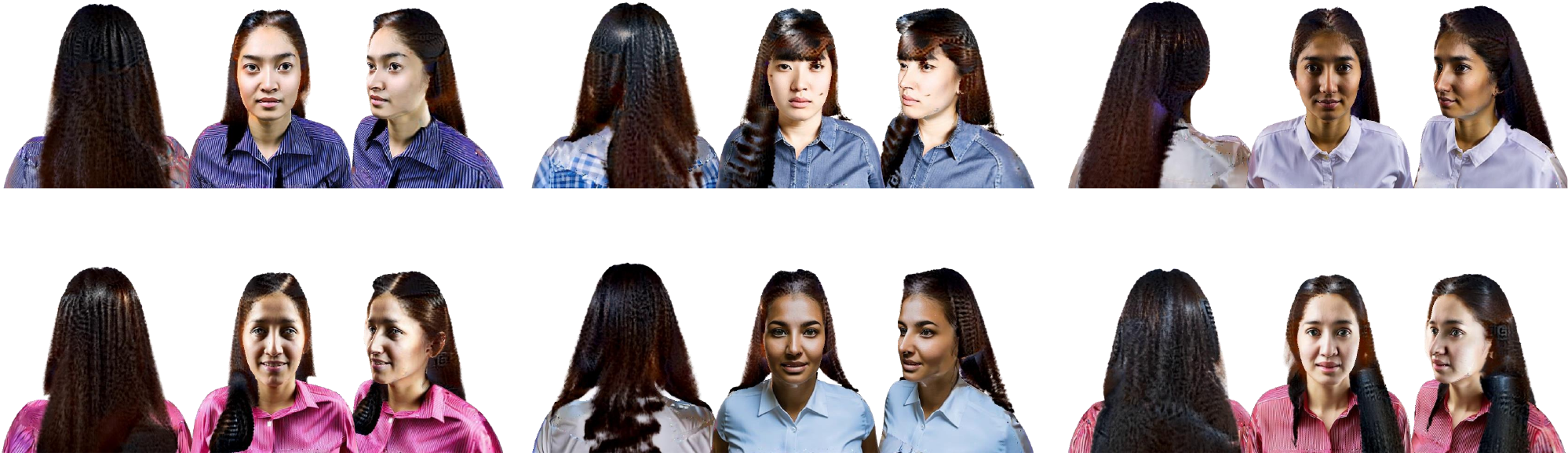}  
 \vspace{-1em}
	\caption{Additional examples of the diversity of avatars generated with the same text prompt: ``a woman with long hair wearing a shirt.''}
	\label{fig:same_seed_diversity}
\end{figure*}

\begin{figure*}[t]
	\centering
	\includegraphics[width=\textwidth, trim={0cm, 0cm, 0cm, 0cm}]{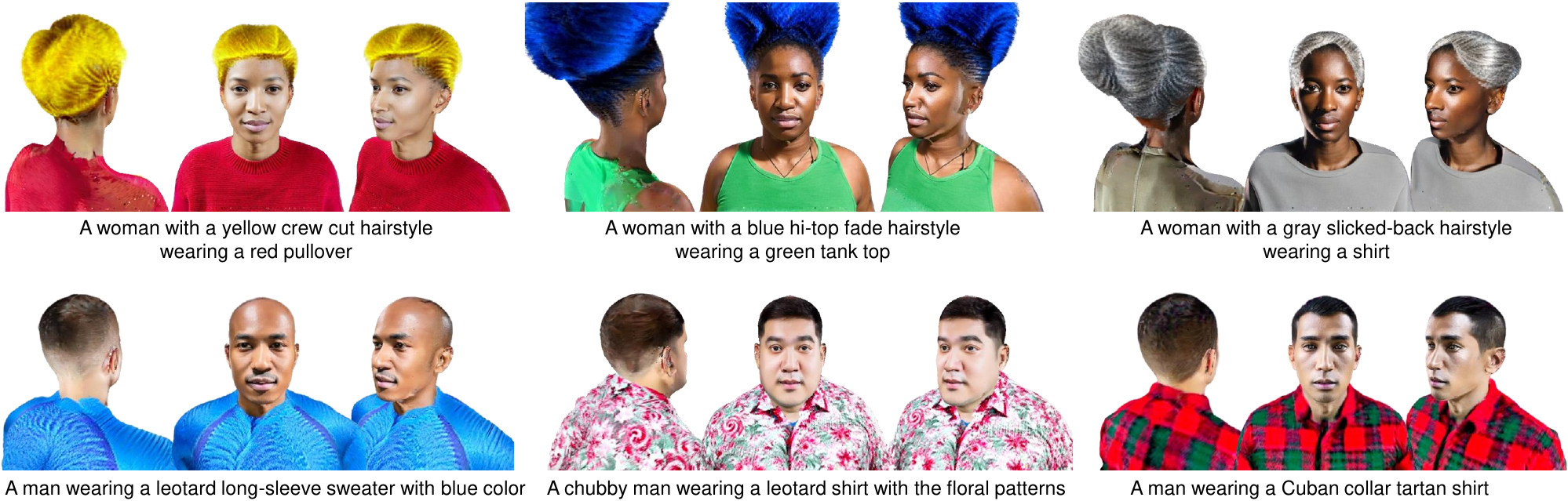}  
 \vspace{-1em}
	\caption{Additional examples of generated avatars with more detailed text descriptions such as colors and detailed shapes.}
	\label{fig:detailed_description}
\end{figure*}

\begin{figure*}[t]
	\centering
	\includegraphics[width=\textwidth, trim={0cm, 0cm, 0cm, 0cm}]{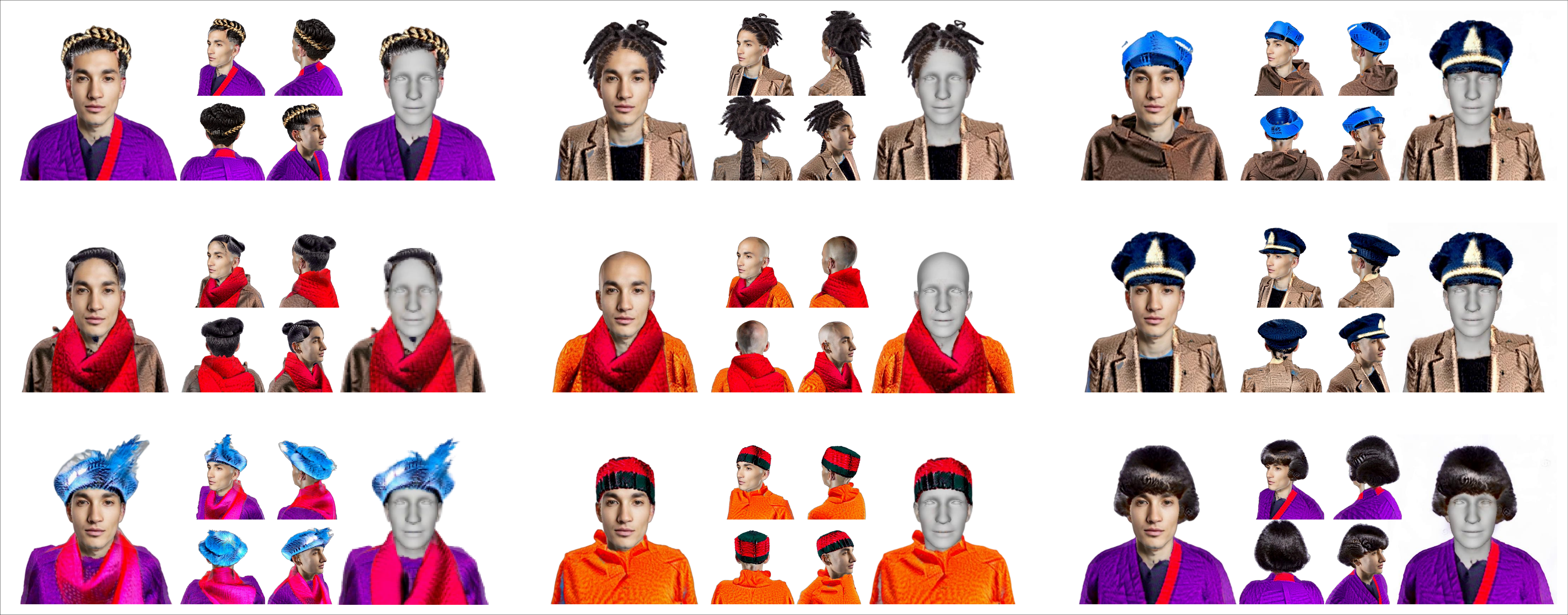}  
 \vspace{-1em}
	\caption{Additional examples for hair and accessory transfer.}
	\label{fig:more_transfer}
\end{figure*}

\begin{figure*}[t]
	\centering
	\includegraphics[width=\textwidth, trim={0cm, 0cm, 0cm, 0cm}]{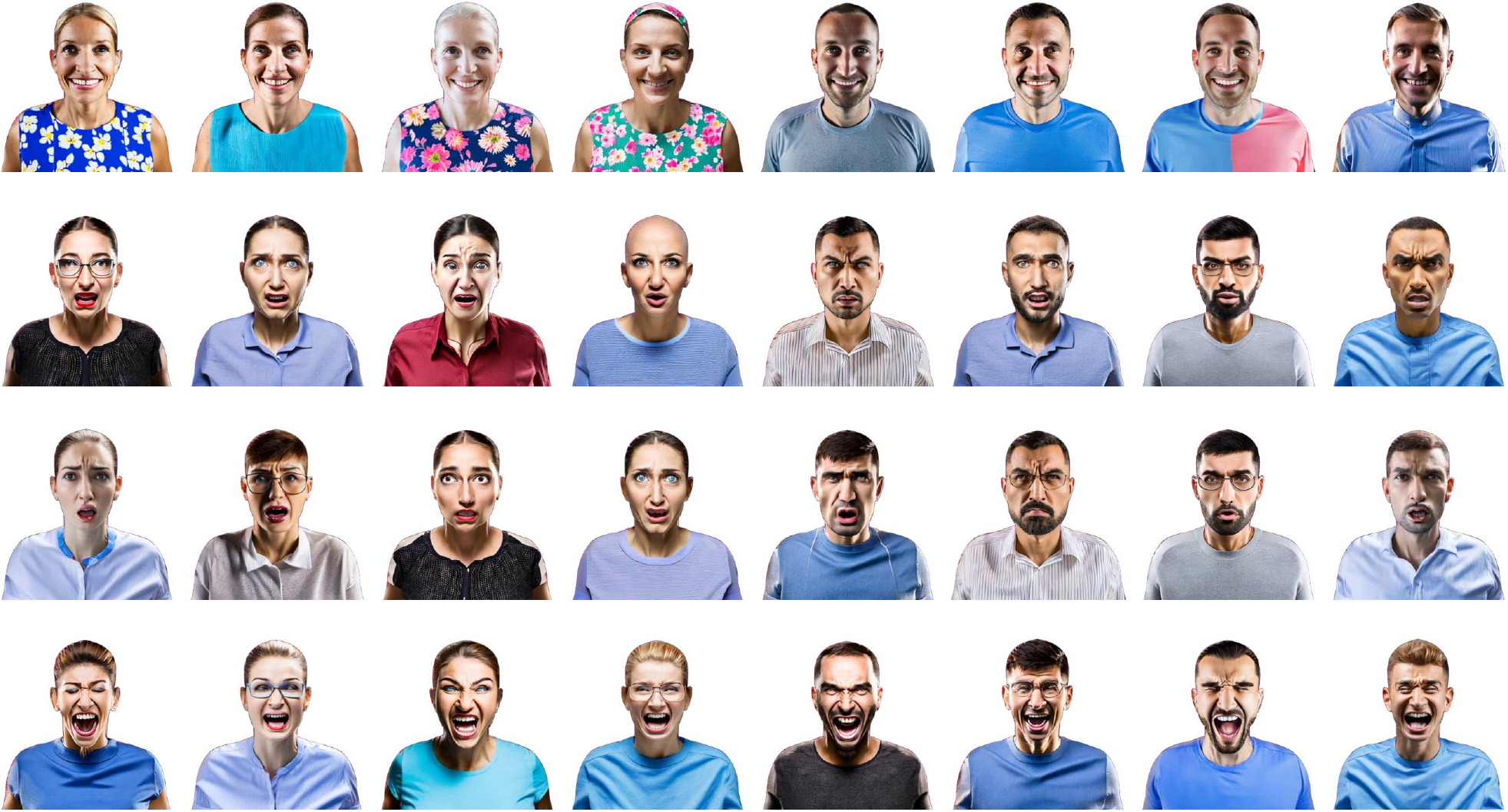}  
 \vspace{-1em}
	\caption{Additional examples avatars generated with different expressions. The in-mouth area is represented with a painted texture. The expressions from top to bottom are: smiling, angry, disgusted, and screaming.}
	\label{fig:different_exp}
 \vspace{-2em}
\end{figure*}

\begin{figure*}[t]
	\centering
	\includegraphics[width=\textwidth, trim={0cm, 0cm, 0cm, 0cm}]{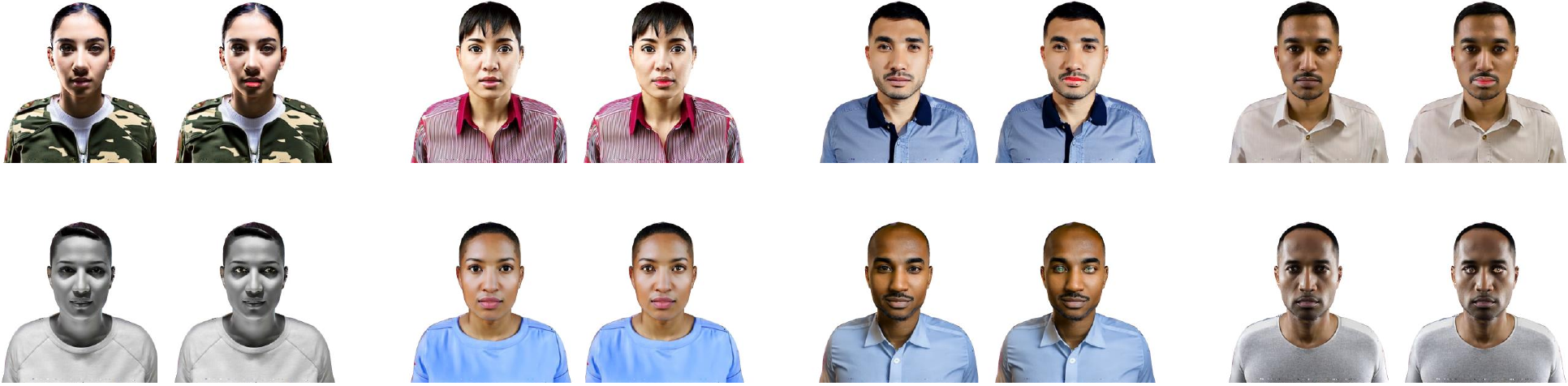}  
 \vspace{-1em}
	\caption{Additional examples of editing the generated avatar's texture. The first line changes the lip color with the ``red lip'' prompt and the second line alters the eye color with prompts ``green eyes'' and ``brown eyes''.}
	\label{fig:edit_texture}

\end{figure*}

\subsection{More Qualitative Results}
\vspace{1em}
\qheading{More generation results.} 
Avatars with distinct hairstyles, clothing, and accessories are shown in Fig.\ \ref{fig:more}. In addition, Fig.\ \ref{fig:more_transfer} shows instances with a variety of other types of accessories, such as earrings, necklaces, and glasses.
To illustrate the diversity of the generated avatars, Fig.\ \ref{fig:same_seed_diversity} shows avatars generated using the same prompt but different seeds, producing diverse faces and clothing that consistently align with the text input. 
Fig.\ \ref{fig:detailed_description} presents avatars that are generated using more detailed descriptions, highlighting the alignment between the avatars and the input texts.

\qheading{More applications.} 
\tdv{
\modelname allows for the transfer of hairstyles and accessories, as demonstrated by additional examples in Fig.\ \ref{fig:more_transfer}.
Moreover, the generated avatars can be animated with different expressions. 
As our method automatically estimates the SMPL-X shape parameters of this subject, 
we can then change the expression parameters of SMPL-X model to animate the face. 
Regarding face texture, the inner mouth region is missing. To address this, we apply an inpainting Stable Diffusion model to inpaint the missing area. 
The results are shown in Fig.\ \ref{fig:different_exp}. 
}
We can further edit the avatar using text input. For example, Fig.\ \ref{fig:edit_texture} shows the results of altering the color of lips and eyes.

\clearpage
\clearpage
{
    \small
    \bibliographystyle{ieeenat_fullname}
    \bibliography{main}

\begin{thebibliography}{71}
\providecommand{\natexlab}[1]{#1}
\providecommand{\url}[1]{\texttt{#1}}
\expandafter\ifx\csname urlstyle\endcsname\relax
  \providecommand{\doi}[1]{doi: #1}\else
  \providecommand{\doi}{doi: \begingroup \urlstyle{rm}\Url}\fi

\bibitem[Alexander et~al.(2010)Alexander, Rogers, Lambeth, Chiang, Ma, Wang,
  and Debevec]{alexander2010digital}
Oleg Alexander, Mike Rogers, William Lambeth, Jen-Yuan Chiang, Wan-Chun Ma,
  Chuan-Chang Wang, and Paul Debevec.
\newblock The {D}igital {E}mily project: {A}chieving a photorealistic digital
  actor.
\newblock \emph{IEEE Computer Graphics and Applications}, 30\penalty0
  (4):\penalty0 20--31, 2010.

\bibitem[Alexander et~al.(2013)Alexander, Fyffe, Busch, Yu, Ichikari, Jones,
  Debevec, Jimenez, Danvoye, Antionazzi, Eheler, Kysela, and von~der
  Pahlen]{alexander2013digital}
Oleg Alexander, Graham Fyffe, Jay Busch, Xueming Yu, Ryosuke Ichikari, Andrew
  Jones, Paul Debevec, Jorge Jimenez, Etienne Danvoye, Bernardo Antionazzi,
  Mike Eheler, Zybnek Kysela, and Javier von~der Pahlen.
\newblock Digital {I}ra: {C}reating a real-time photoreal digital actor.
\newblock In \emph{ACM SIGGRAPH 2013 Posters}, New York, NY, USA, 2013.
  Association for Computing Machinery.

\bibitem[An et~al.(2023)An, Xu, Shi, Song, Ogras, and Luo]{An2023PanoHeadG3}
Sizhe An, Hongyi Xu, Yichun Shi, Guoxian Song, Umit Ogras, and Linjie Luo.
\newblock Panohead: Geometry-aware 3d full-head synthesis in 360$^{\circ}$,
  2023.

\bibitem[Aneja et~al.(2023)Aneja, Thies, Dai, and Nießner]{aneja2022clipface}
Shivangi Aneja, Justus Thies, Angela Dai, and Matthias Nießner.
\newblock Clipface: Text-guided editing of textured 3d morphable models, 2023.

\bibitem[Blanz and Vetter(1999)]{blanz1999morphable}
Volker Blanz and Thomas Vetter.
\newblock A morphable model for the synthesis of 3d faces.
\newblock In \emph{Transactions on Graphics, (Proc. SIGGRAPH)}, page 187–194,
  USA, 1999. ACM Press/Addison-Wesley Publishing Co.

\bibitem[Borshukov and Lewis(2005)]{borshukov2005realistic}
George Borshukov and J.~P. Lewis.
\newblock Realistic human face rendering for "the matrix reloaded".
\newblock In \emph{ACM SIGGRAPH 2005 Courses}, page 13–es, New York, NY, USA,
  2005. Association for Computing Machinery.

\bibitem[Bulat and Tzimiropoulos(2017)]{bulat2017far}
A. Bulat and G. Tzimiropoulos.
\newblock How far are we from solving the 2d \& 3d face alignment problem? (and
  a dataset of 230,000 3d facial landmarks).
\newblock In \emph{2017 IEEE International Conference on Computer Vision
  (ICCV)}, pages 1021--1030, Los Alamitos, CA, USA, 2017. IEEE Computer
  Society.

\bibitem[Cao et~al.(2022)Cao, Simon, Kim, Schwartz, Zollhoefer, Saito,
  Lombardi, Wei, Belko, Yu, et~al.]{cao2022authentic}
Chen Cao, Tomas Simon, Jin~Kyu Kim, Gabe Schwartz, Michael Zollhoefer,
  Shun-Suke Saito, Stephen Lombardi, Shih-En Wei, Danielle Belko, Shoou-I Yu,
  et~al.
\newblock Authentic volumetric avatars from a phone scan.
\newblock \emph{ACM Transactions on Graphics (TOG)}, 41\penalty0 (4):\penalty0
  1--19, 2022.

\bibitem[Cao et~al.(2023)Cao, Cao, Han, Shan, and Wong]{cao2023dreamavatar}
Yukang Cao, Yan-Pei Cao, Kai Han, Ying Shan, and Kwan-Yee~K. Wong.
\newblock Dreamavatar: Text-and-shape guided 3d human avatar generation via
  diffusion models, 2023.

\bibitem[Chan et~al.(2022)Chan, Lin, Chan, Nagano, Pan, De~Mello, Gallo,
  Guibas, Tremblay, Khamis, Karras, and Wetzstein]{Chan2021eg3d}
Eric~R. Chan, Connor~Z. Lin, Matthew~A. Chan, Koki Nagano, Boxiao Pan, Shalini
  De~Mello, Orazio Gallo, Leonidas~J. Guibas, Jonathan Tremblay, Sameh Khamis,
  Tero Karras, and Gordon Wetzstein.
\newblock Efficient geometry-aware 3d generative adversarial networks.
\newblock In \emph{Conference on Computer Vision and Pattern Recognition
  (CVPR)}, pages 16123--16133, Los Alamitos, CA, USA, 2022. IEEE Computer
  Society.

\bibitem[Chen et~al.(2021)Chen, Zhang, Kang, Zhe, Bao, Jia, and
  Lu]{chen2021animatable}
Jianchuan Chen, Ying Zhang, Di Kang, Xuefei Zhe, Linchao Bao, Xu Jia, and
  Huchuan Lu.
\newblock Animatable neural radiance fields from monocular rgb videos, 2021.

\bibitem[Chen et~al.(2023)Chen, Chen, Jiao, and Jia]{chen2023fantasia3d}
Rui Chen, Yongwei Chen, Ningxin Jiao, and Kui Jia.
\newblock Fantasia3d: Disentangling geometry and appearance for high-quality
  text-to-3d content creation, 2023.

\bibitem[Chen et~al.(2022)Chen, Chen, Lei, Zhang, and Jia]{chen2022tango}
Yongwei Chen, Rui Chen, Jiabao Lei, Yabin Zhang, and Kui Jia.
\newblock Tango: Text-driven photorealistic and robust 3d stylization via
  lighting decomposition, 2022.

\bibitem[{Epic Games}(2023)]{metahuman}
{Epic Games}.
\newblock Metahuman, 2023.

\bibitem[Feng et~al.(2018)Feng, Wu, Shao, Wang, and Zhou]{feng2018prn}
Yao Feng, Fan Wu, Xiaohu Shao, Yanfeng Wang, and Xi Zhou.
\newblock Joint 3d face reconstruction and dense alignment with position map
  regression network.
\newblock In \emph{ECCV}, 2018.

\bibitem[Feng et~al.(2021{\natexlab{a}})Feng, Choutas, Bolkart, Tzionas, and
  Black]{PIXIE:3DV:2021}
Yao Feng, Vasileios Choutas, Timo Bolkart, Dimitrios Tzionas, and Michael
  Black.
\newblock Collaborative regression of expressive bodies using moderation.
\newblock In \emph{International Conference on 3D Vision (3DV)}, pages
  792--804, 2021{\natexlab{a}}.

\bibitem[Feng et~al.(2021{\natexlab{b}})Feng, Feng, Black, and
  Bolkart]{feng2021learning}
Yao Feng, Haiwen Feng, Michael~J. Black, and Timo Bolkart.
\newblock Learning an animatable detailed {3D} face model from in-the-wild
  images.
\newblock \emph{Transactions on Graphics, (Proc. SIGGRAPH)}, 40\penalty0
  (4):\penalty0 1--13, 2021{\natexlab{b}}.

\bibitem[Feng et~al.(2022)Feng, Yang, Pollefeys, Black, and
  Bolkart]{feng2022capturing}
Yao Feng, Jinlong Yang, Marc Pollefeys, Michael~J. Black, and Timo Bolkart.
\newblock Capturing and animation of body and clothing from monocular video.
\newblock In \emph{SIGGRAPH Asia 2022 Conference Papers}, New York, NY, USA,
  2022. Association for Computing Machinery.

\bibitem[Feng et~al.(2023)Feng, Liu, Bolkart, Yang, Pollefeys, and
  Black]{Feng2023DELTA}
Yao Feng, Weiyang Liu, Timo Bolkart, Jinlong Yang, Marc Pollefeys, and
  Michael~J. Black.
\newblock Learning disentangled avatars with hybrid 3d representations.
\newblock \emph{arXiv}, 2023.

\bibitem[Gao et~al.(2022)Gao, Zhong, Xiang, Hong, Guo, and
  Zhang]{gao2022reconstructing}
Xuan Gao, Chenglai Zhong, Jun Xiang, Yang Hong, Yudong Guo, and Juyong Zhang.
\newblock Reconstructing personalized semantic facial {NeRF} models from
  monocular video.
\newblock \emph{Transactions on Graphics (TOG)}, 41\penalty0 (6):\penalty0
  1--12, 2022.

\bibitem[Gerig et~al.(2018)Gerig, Morel-Forster, Blumer, Egger, Luthi,
  Schoenborn, and Vetter]{basel}
Thomas Gerig, Andreas Morel-Forster, Clemens Blumer, Bernhard Egger, Marcel
  Luthi, Sandro Schoenborn, and Thomas Vetter.
\newblock Morphable face models - an open framework.
\newblock In \emph{2018 13th IEEE International Conference on Automatic Face \&
  Gesture Recognition (FG 2018)}, pages 75--82, Los Alamitos, CA, USA, 2018.
  IEEE Computer Society.

\bibitem[Grassal et~al.(2022)Grassal, Prinzler, Leistner, Rother, Nie{\ss}ner,
  and Thies]{grassal2022neural}
Philip-William Grassal, Malte Prinzler, Titus Leistner, Carsten Rother,
  Matthias Nie{\ss}ner, and Justus Thies.
\newblock Neural head avatars from monocular rgb videos.
\newblock In \emph{Conference on Computer Vision and Pattern Recognition
  (CVPR)}, pages 18653--18664, Los Alamitos, CA, USA, 2022. IEEE Computer
  Society.

\bibitem[Guo et~al.(2019)Guo, Lincoln, Davidson, Busch, Yu, Whalen, Harvey,
  Orts-Escolano, Pandey, Dourgarian, et~al.]{guo2019relightables}
Kaiwen Guo, Peter Lincoln, Philip Davidson, Jay Busch, Xueming Yu, Matt Whalen,
  Geoff Harvey, Sergio Orts-Escolano, Rohit Pandey, Jason Dourgarian, et~al.
\newblock The relightables: Volumetric performance capture of humans with
  realistic relighting.
\newblock \emph{Transactions on Graphics (TOG)}, 38\penalty0 (6):\penalty0
  1--19, 2019.

\bibitem[Ho et~al.(2020)Ho, Jain, and Abbeel]{ho2020denoising}
Jonathan Ho, Ajay Jain, and Pieter Abbeel.
\newblock Denoising diffusion probabilistic models.
\newblock \emph{Advances in Neural Information Processing Systems (NeurIPS)},
  33:\penalty0 6840--6851, 2020.

\bibitem[Hong et~al.(2022)Hong, Zhang, Pan, Cai, Yang, and
  Liu]{hong2022avatarclip}
Fangzhou Hong, Mingyuan Zhang, Liang Pan, Zhongang Cai, Lei Yang, and Ziwei
  Liu.
\newblock {AvatarCLIP}: {Z}ero-shot text-driven generation and animation of
  {3D} avatars.
\newblock \emph{Transactions on Graphics (TOG)}, 41\penalty0 (4):\penalty0
  1--19, 2022.

\bibitem[Huang et~al.(2023)Huang, Xiu, Yi, Liao, Tang, Cai, and
  Thies]{huang2023tech}
Yangyi Huang, Yuliang Xiu, Hongwei Yi, Tingting Liao, Jiaxiang Tang, Deng Cai,
  and Justus Thies.
\newblock {TeCH: Text-guided Reconstruction of Lifelike Clothed Humans}.
\newblock \emph{arXiv preprint: 2308.08545}, 2023.

\bibitem[Jain et~al.(2022)Jain, Mildenhall, Barron, Abbeel, and
  Poole]{jain2022zero}
Ajay Jain, Ben Mildenhall, Jonathan~T. Barron, Pieter Abbeel, and Ben Poole.
\newblock Zero-shot text-guided object generation with dream fields.
\newblock In \emph{Conference on Computer Vision and Pattern Recognition
  (CVPR)}, pages 867--876, Los Alamitos, CA, USA, 2022. IEEE Computer Society.

\bibitem[Karras et~al.(2019)Karras, Laine, and Aila]{karras2019style}
Tero Karras, Samuli Laine, and Timo Aila.
\newblock A style-based generator architecture for generative adversarial
  networks.
\newblock In \emph{Proceedings of the IEEE/CVF conference on computer vision
  and pattern recognition}, pages 4401--4410, 2019.

\bibitem[Kingma and Ba(2014)]{kingma2014adam}
Diederik~P Kingma and Jimmy Ba.
\newblock Adam: A method for stochastic optimization.
\newblock \emph{arXiv preprint arXiv:1412.6980}, 2014.

\bibitem[Li et~al.(2023{\natexlab{a}})Li, Li, Le, Wang, Savarese, and
  Hoi]{li-etal-2023-lavis}
Dongxu Li, Junnan Li, Hung Le, Guangsen Wang, Silvio Savarese, and Steven~C.H.
  Hoi.
\newblock {LAVIS}: A one-stop library for language-vision intelligence.
\newblock In \emph{Proceedings of the 61st Annual Meeting of the Association
  for Computational Linguistics (Volume 3: System Demonstrations)}, pages
  31--41, Toronto, Canada, 2023{\natexlab{a}}. Association for Computational
  Linguistics.

\bibitem[Li et~al.(2022)Li, Li, Xiong, and Hoi]{li2022blip}
Junnan Li, Dongxu Li, Caiming Xiong, and Steven Hoi.
\newblock {BLIP}: Bootstrapping language-image pre-training for unified
  vision-language understanding and generation.
\newblock In \emph{International Conference on Machine Learning (ICML)}, pages
  12888--12900, Baltimore, MD, USA, 2022. PMLR.

\bibitem[Li et~al.(2023{\natexlab{b}})Li, Saito, Simon, Lombardi, Li, and
  Saragih]{li2023megane}
Junxuan Li, Shunsuke Saito, Tomas Simon, Stephen Lombardi, Hongdong Li, and
  Jason Saragih.
\newblock Megane: Morphable eyeglass and avatar network, 2023{\natexlab{b}}.

\bibitem[Li et~al.(2017{\natexlab{a}})Li, Bolkart, Black, Li, and
  Romero]{FLAME:SiggraphAsia2017}
Tianye Li, Timo Bolkart, Michael.~J. Black, Hao Li, and Javier Romero.
\newblock Learning a model of facial shape and expression from {4D} scans.
\newblock \emph{ACM Transactions on Graphics, (Proc. SIGGRAPH Asia)},
  36\penalty0 (6):\penalty0 194:1--194:17, 2017{\natexlab{a}}.

\bibitem[Li et~al.(2017{\natexlab{b}})Li, Bolkart, Black, Li, and
  Romero]{li2017learning}
Tianye Li, Timo Bolkart, Michael~J. Black, Hao Li, and Javier Romero.
\newblock Learning a model of facial shape and expression from {4D} scans.
\newblock \emph{Transactions on Graphics (TOG)}, 36\penalty0 (6):\penalty0
  1--17, 2017{\natexlab{b}}.

\bibitem[Lin et~al.(2023)Lin, Gao, Tang, Takikawa, Zeng, Huang, Kreis, Fidler,
  Liu, and Lin]{lin2022magic3d}
Chen-Hsuan Lin, Jun Gao, Luming Tang, Towaki Takikawa, Xiaohui Zeng, Xun Huang,
  Karsten Kreis, Sanja Fidler, Ming-Yu Liu, and Tsung-Yi Lin.
\newblock Magic3d: High-resolution text-to-3d content creation, 2023.

\bibitem[Liu et~al.(2022)Liu, Liu, Paull, Weller, and Schölkopf]{Liu2022SCR}
Weiyang Liu, Zhen Liu, Liam Paull, Adrian Weller, and Bernhard Schölkopf.
\newblock Structural causal 3d reconstruction.
\newblock In \emph{ECCV}, 2022.

\bibitem[Liu et~al.(2015)Liu, Luo, Wang, and Tang]{liu2015deep}
Z. Liu, P. Luo, X. Wang, and X. Tang.
\newblock Deep learning face attributes in the wild.
\newblock In \emph{2015 IEEE International Conference on Computer Vision
  (ICCV)}, pages 3730--3738, Los Alamitos, CA, USA, 2015. IEEE Computer
  Society.

\bibitem[Lombardi et~al.(2018)Lombardi, Saragih, Simon, and
  Sheikh]{lombardi2018deep}
Stephen Lombardi, Jason Saragih, Tomas Simon, and Yaser Sheikh.
\newblock Deep appearance models for face rendering.
\newblock \emph{Transactions on Graphics (TOG)}, 37\penalty0 (4):\penalty0
  1--13, 2018.

\bibitem[Loper et~al.(2015{\natexlab{a}})Loper, Mahmood, Romero, Pons-Moll, and
  Black]{SMPL:2015}
Matthew Loper, Naureen Mahmood, Javier Romero, Gerard Pons-Moll, and Michael~J.
  Black.
\newblock {SMPL}: {A} skinned multi-person linear model.
\newblock \emph{ACM Transactions on Graphics, (Proc. SIGGRAPH Asia)},
  34\penalty0 (6):\penalty0 248:1--248:16, 2015{\natexlab{a}}.

\bibitem[Loper et~al.(2015{\natexlab{b}})Loper, Mahmood, Romero, Pons-Moll, and
  Black]{loper2015smpl}
Matthew Loper, Naureen Mahmood, Javier Romero, Gerard Pons-Moll, and Michael~J.
  Black.
\newblock {SMPL}: {A} skinned multi-person linear model.
\newblock \emph{Transactions on Graphics (TOG)}, 34\penalty0 (6):\penalty0
  1--16, 2015{\natexlab{b}}.

\bibitem[L\"uddecke and Ecker(2022)]{lueddecke22_cvpr}
Timo L\"uddecke and Alexander Ecker.
\newblock Image segmentation using text and image prompts.
\newblock In \emph{Conference on Computer Vision and Pattern Recognition
  (CVPR)}, pages 7086--7096, Los Alamitos, CA, USA, 2022. IEEE Computer
  Society.

\bibitem[Metzer et~al.(2022)Metzer, Richardson, Patashnik, Giryes, and
  Cohen-Or]{metzer2022latent}
Gal Metzer, Elad Richardson, Or Patashnik, Raja Giryes, and Daniel Cohen-Or.
\newblock Latent-nerf for shape-guided generation of 3d shapes and textures,
  2022.

\bibitem[Michel et~al.(2022)Michel, Bar-On, Liu, Benaim, and
  Hanocka]{michel2022text2mesh}
Oscar Michel, Roi Bar-On, Richard Liu, Sagie Benaim, and Rana Hanocka.
\newblock Text2mesh: Text-driven neural stylization for meshes.
\newblock In \emph{Conference on Computer Vision and Pattern Recognition
  (CVPR)}, pages 13492--13502, Los Alamitos, CA, USA, 2022. IEEE Computer
  Society.

\bibitem[Mildenhall et~al.(2021)Mildenhall, Srinivasan, Tancik, Barron,
  Ramamoorthi, and Ng]{mildenhall2020nerf}
Ben Mildenhall, Pratul~P. Srinivasan, Matthew Tancik, Jonathan~T. Barron, Ravi
  Ramamoorthi, and Ren Ng.
\newblock Nerf: Representing scenes as neural radiance fields for view
  synthesis.
\newblock \emph{Commun. ACM}, 65\penalty0 (1):\penalty0 99–106, 2021.

\bibitem[Mohammad~Khalid et~al.(2022)Mohammad~Khalid, Xie, Belilovsky, and
  Popa]{mohammad2022clip}
Nasir Mohammad~Khalid, Tianhao Xie, Eugene Belilovsky, and Tiberiu Popa.
\newblock Clip-mesh: Generating textured meshes from text using pretrained
  image-text models.
\newblock In \emph{SIGGRAPH Asia 2022 Conference Papers}, New York, NY, USA,
  2022. Association for Computing Machinery.

\bibitem[Nichol et~al.(2022)Nichol, Dhariwal, Ramesh, Shyam, Mishkin, Mcgrew,
  Sutskever, and Chen]{nichol2021glide}
Alexander~Quinn Nichol, Prafulla Dhariwal, Aditya Ramesh, Pranav Shyam, Pamela
  Mishkin, Bob Mcgrew, Ilya Sutskever, and Mark Chen.
\newblock {GLIDE}: Towards photorealistic image generation and editing with
  text-guided diffusion models.
\newblock In \emph{International Conference on Machine Learning (ICML)}, pages
  16784--16804, Virtual, 2022. PMLR.

\bibitem[Pavlakos et~al.(2019)Pavlakos, Choutas, Ghorbani, Bolkart, Osman,
  Tzionas, and Black]{Pavlakos2019_smplifyx}
G. Pavlakos, V. Choutas, N. Ghorbani, T. Bolkart, A.~A. Osman, D. Tzionas, and
  M.~J. Black.
\newblock Expressive body capture: {3D} hands, face, and body from a single
  image.
\newblock In \emph{2019 IEEE/CVF Conference on Computer Vision and Pattern
  Recognition (CVPR)}, pages 10967--10977, Los Alamitos, CA, USA, 2019. IEEE
  Computer Society.

\bibitem[Poole et~al.(2022)Poole, Jain, Barron, and
  Mildenhall]{poole2023dreamfusion}
Ben Poole, Ajay Jain, Jonathan~T. Barron, and Ben Mildenhall.
\newblock Dreamfusion: Text-to-3d using 2d diffusion, 2022.

\bibitem[Radford et~al.(2021)Radford, Kim, Hallacy, Ramesh, Goh, Agarwal,
  Sastry, Askell, Mishkin, Clark, Krueger, and Sutskever]{radford2021learning}
Alec Radford, Jong~Wook Kim, Chris Hallacy, Aditya Ramesh, Gabriel Goh,
  Sandhini Agarwal, Girish Sastry, Amanda Askell, Pamela Mishkin, Jack Clark,
  Gretchen Krueger, and Ilya Sutskever.
\newblock Learning transferable visual models from natural language
  supervision.
\newblock In \emph{International Conference on Machine Learning (ICML)}, pages
  8748--8763, Virtual, 2021. PMLR.

\bibitem[Ramesh et~al.(2022)Ramesh, Dhariwal, Nichol, Chu, and
  Chen]{ramesh2022hierarchical}
Aditya Ramesh, Prafulla Dhariwal, Alex Nichol, Casey Chu, and Mark Chen.
\newblock Hierarchical text-conditional image generation with clip latents,
  2022.

\bibitem[Ranade et~al.(2022)Ranade, Lassner, Li, Haene, Chen, Bazin, and
  Bouaziz]{ranade2022ssdnerf}
Siddhant Ranade, Christoph Lassner, Kai Li, Christian Haene, Shen-Chi Chen,
  Jean-Charles Bazin, and Sofien Bouaziz.
\newblock Ssdnerf: Semantic soft decomposition of neural radiance fields, 2022.

\bibitem[Richardson et~al.(2023)Richardson, Metzer, Alaluf, Giryes, and
  Cohen-Or]{richardson2023texture}
Elad Richardson, Gal Metzer, Yuval Alaluf, Raja Giryes, and Daniel Cohen-Or.
\newblock Texture: Text-guided texturing of 3d shapes, 2023.

\bibitem[Rombach et~al.(2022)Rombach, Blattmann, Lorenz, Esser, and
  Ommer]{rombach2022high}
Robin Rombach, Andreas Blattmann, Dominik Lorenz, Patrick Esser, and Bj\"orn
  Ommer.
\newblock High-resolution image synthesis with latent diffusion models.
\newblock In \emph{Conference on Computer Vision and Pattern Recognition
  (CVPR)}, pages 10684--10695, Los Alamitos, CA, USA, 2022. IEEE Computer
  Society.

\bibitem[Rosu et~al.(2022)Rosu, Saito, Wang, Wu, Behnke, and
  Nam]{rosu2022neural}
Radu~Alexandru Rosu, Shunsuke Saito, Ziyan Wang, Chenglei Wu, Sven Behnke, and
  Giljoo Nam.
\newblock Neural strands: Learning hair geometry and appearance from multi-view
  images.
\newblock In \emph{Computer Vision -- ECCV 2022}, pages 73--89, Cham, 2022.
  Springer Nature Switzerland.

\bibitem[Sanghi et~al.(2022)Sanghi, Chu, Lambourne, Wang, Cheng, Fumero, and
  Malekshan]{sanghi2022clip}
Aditya Sanghi, Hang Chu, Joseph~G. Lambourne, Ye Wang, Chin-Yi Cheng, Marco
  Fumero, and Kamal~Rahimi Malekshan.
\newblock Clip-forge: Towards zero-shot text-to-shape generation.
\newblock In \emph{Conference on Computer Vision and Pattern Recognition
  (CVPR)}, pages 18603--18613, Los Alamitos, CA, USA, 2022. IEEE Computer
  Society.

\bibitem[Seitzer(2020)]{Seitzer2020FID}
Maximilian Seitzer.
\newblock {pytorch-fid: FID Score for PyTorch}.
\newblock \url{https://github.com/mseitzer/pytorch-fid}, 2020.
\newblock Version 0.3.0.

\bibitem[Seymour et~al.(2017)Seymour, Evans, and Libreri]{seymour2017meet}
Mike Seymour, Chris Evans, and Kim Libreri.
\newblock Meet mike: Epic avatars.
\newblock In \emph{ACM SIGGRAPH 2017 VR Village}, New York, NY, USA, 2017.
  Association for Computing Machinery.

\bibitem[Shao et~al.(2023)Shao, Sun, Peng, Zheng, Zhou, Zhang, and
  Liu]{shao2023control4d}
Ruizhi Shao, Jingxiang Sun, Cheng Peng, Zerong Zheng, Boyao Zhou, Hongwen
  Zhang, and Yebin Liu.
\newblock Control4d: Dynamic portrait editing by learning 4d gan from 2d
  diffusion-based editor.
\newblock \emph{arXiv preprint arXiv:2305.20082}, 2023.

\bibitem[Tewari et~al.(2017)Tewari, Zollhöfer, Kim, Garrido, Bernard, Pérez,
  and Theobalt]{tewari2017mofa}
Ayush Tewari, Michael Zollhöfer, Hyeongwoo Kim, Pablo Garrido, Florian
  Bernard, Patrick Pérez, and Christian Theobalt.
\newblock Mofa: Model-based deep convolutional face autoencoder for
  unsupervised monocular reconstruction.
\newblock In \emph{International Conference on Computer Vision (ICCV)}, pages
  3735--3744, Los Alamitos, CA, USA, 2017. IEEE Computer Society.

\bibitem[Turner(2022)]{KevinTurner2022}
Kevin Turner.
\newblock Decoding latents to {RGB} without upscaling, 2022.

\bibitem[Wang et~al.(2022)Wang, Chai, He, Chen, and Liao]{wang2022clip}
Can Wang, Menglei Chai, Mingming He, Dongdong Chen, and Jing Liao.
\newblock {CLIP-NeRF}: Text-and-image driven manipulation of neural radiance
  fields.
\newblock In \emph{Conference on Computer Vision and Pattern Recognition
  (CVPR)}, pages 3835--3844, Los Alamitos, CA, USA, 2022. IEEE Computer
  Society.

\bibitem[Wang et~al.(2023{\natexlab{a}})Wang, Du, Li, Yeh, and
  Shakhnarovich]{wang2022score}
Haochen Wang, Xiaodan Du, Jiahao Li, Raymond~A. Yeh, and Greg Shakhnarovich.
\newblock Score jacobian chaining: Lifting pretrained 2d diffusion models for
  3d generation.
\newblock In \emph{Conference on Computer Vision and Pattern Recognition
  (CVPR)}, pages 12619--12629, Los Alamitos, CA, USA, 2023{\natexlab{a}}. IEEE
  Computer Society.

\bibitem[Wang et~al.(2023{\natexlab{b}})Wang, Zhang, Zhang, Gu, Bao,
  Baltrusaitis, Shen, Chen, Wen, Chen, and Guo]{wang2022rodin}
Tengfei Wang, Bo Zhang, Ting Zhang, Shuyang Gu, Jianmin Bao, Tadas
  Baltrusaitis, Jingjing Shen, Dong Chen, Fang Wen, Qifeng Chen, and Baining
  Guo.
\newblock Rodin: A generative model for sculpting 3d digital avatars using
  diffusion.
\newblock In \emph{Conference on Computer Vision and Pattern Recognition
  (CVPR)}, pages 4563--4573, Los Alamitos, CA, USA, 2023{\natexlab{b}}. IEEE
  Computer Society.

\bibitem[Wang et~al.(2023{\natexlab{c}})Wang, Lu, Wang, Bao, Li, Su, and
  Zhu]{wang2023prolificdreamer}
Zhengyi Wang, Cheng Lu, Yikai Wang, Fan Bao, Chongxuan Li, Hang Su, and Jun
  Zhu.
\newblock Prolificdreamer: High-fidelity and diverse text-to-3d generation with
  variational score distillation.
\newblock \emph{arXiv preprint arXiv:2305.16213}, 2023{\natexlab{c}}.

\bibitem[Wei et~al.(2022)Wei, Chen, Zhou, Liao, Tan, Yuan, Zhang, and
  Yu]{wei2022hairclip}
Tianyi Wei, Dongdong Chen, Wenbo Zhou, Jing Liao, Zhentao Tan, Lu Yuan, Weiming
  Zhang, and Nenghai Yu.
\newblock Hairclip: Design your hair by text and reference image.
\newblock In \emph{Conference on Computer Vision and Pattern Recognition
  (CVPR)}, pages 18072--18081, Los Alamitos, CA, USA, 2022. IEEE Computer
  Society.

\bibitem[Wu et~al.(2023)Wu, Zhu, Huang, Zhuang, Lu, and Cao]{wu2023describe3d}
Menghua Wu, Hao Zhu, Linjia Huang, Yiyu Zhuang, Yuanxun Lu, and Xun Cao.
\newblock High-fidelity 3d face generation from natural language descriptions.
\newblock In \emph{Conference on Computer Vision and Pattern Recognition
  (CVPR)}, pages 4521--4530, Los Alamitos, CA, USA, 2023. IEEE Computer
  Society.

\bibitem[Zhang et~al.(2023)Zhang, Qiu, Lin, Zhang, Shi, Yang, Shi, Yang, Xu,
  and Yu]{zhang2023dreamface}
Longwen Zhang, Qiwei Qiu, Hongyang Lin, Qixuan Zhang, Cheng Shi, Wei Yang, Ye
  Shi, Sibei Yang, Lan Xu, and Jingyi Yu.
\newblock Dreamface: Progressive generation of animatable 3d faces under text
  guidance, 2023.

\bibitem[Zhao et~al.(2023)Zhao, Li, Hu, Li, Zou, Shi, and Fan]{zhao2023zero}
Rui Zhao, Wei Li, Zhipeng Hu, Lincheng Li, Zhengxia Zou, Zhenwei Shi, and
  Changjie Fan.
\newblock Zero-shot text-to-parameter translation for game character
  auto-creation.
\newblock In \emph{Conference on Computer Vision and Pattern Recognition
  (CVPR)}, pages 21013--21023, Los Alamitos, CA, USA, 2023. IEEE Computer
  Society.

\bibitem[Zheng et~al.(2022)Zheng, Abrevaya, B\"uhler, Chen, Black, and
  Hilliges]{zheng2022avatar}
Yufeng Zheng, Victoria~Fern\'andez Abrevaya, Marcel~C. B\"uhler, Xu Chen,
  Michael~J. Black, and Otmar Hilliges.
\newblock I m avatar: Implicit morphable head avatars from videos.
\newblock In \emph{Conference on Computer Vision and Pattern Recognition
  (CVPR)}, pages 13545--13555, Los Alamitos, CA, USA, 2022. IEEE Computer
  Society.

\bibitem[Zielonka et~al.(2022{\natexlab{a}})Zielonka, Bolkart, and
  Thies]{zielonka2022instant}
Wojciech Zielonka, Timo Bolkart, and Justus Thies.
\newblock Instant volumetric head avatars, 2022{\natexlab{a}}.

\bibitem[Zielonka et~al.(2022{\natexlab{b}})Zielonka, Bolkart, and
  Thies]{zielonka2022towards}
Wojciech Zielonka, Timo Bolkart, and Justus Thies.
\newblock Towards metrical reconstruction of human faces.
\newblock In \emph{Computer Vision -- ECCV 2022}, pages 250--269, Cham,
  2022{\natexlab{b}}. Springer Nature Switzerland.

\end{thebibliography}
}

\end{document}